%% file: 0-main.tex
\documentclass[conference]{IEEEtran}
\usepackage{times}

\usepackage[numbers]{natbib}
\usepackage{multicol}
\usepackage{flushend} %balance refs on last page
\usepackage[bookmarks=true,
            urlcolor=blue,
            citecolor=blue]{hyperref}

\include{0-format_and_definitions}
\newcommand{\algoName}{GIFT\xspace}

\begin{document}

\makeatletter
    \let\@oldmaketitle\@maketitle% Store \@maketitle
    \renewcommand{\@maketitle}{\@oldmaketitle% Update \@maketitle to insert...
    \centering
    \includegraphics[width=\linewidth]{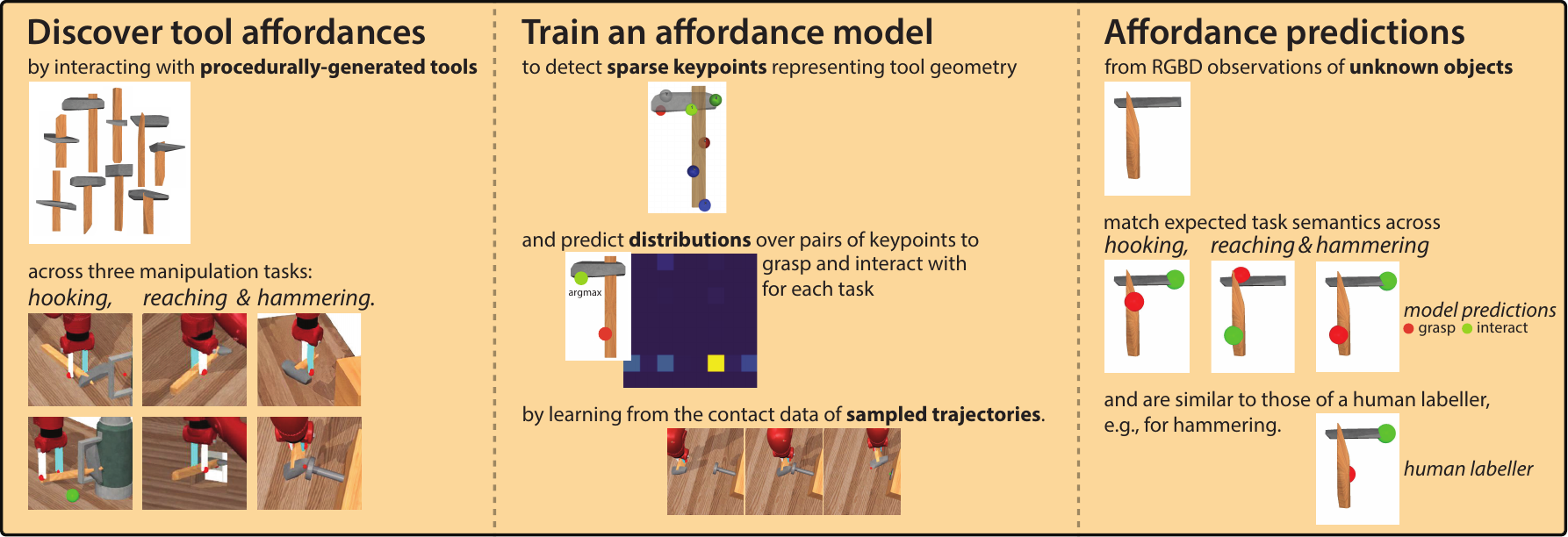}
    \captionof{figure}{Rather than relying on human labels, the \textbf{GIFT} framework discovers affordances from goal-directed
    interaction with a set of procedurally-generated tools.
    This interaction experience is collected with a simple sampling-based motion planner that does not require
    demonstrations or an expert policy.
    Since the affordances are not prespecified (either explicitly by labels or implicitly by predefined manipulation strategies),
    they are unbiased, i.e., they emerge only from the constraints of the task.}
    \label{fig:teaser}
    }
\makeatother

\title{GIFT: Generalizable Interaction-aware Functional Tool Affordances without Labels}

\author{\IEEEauthorblockN{Dylan Turpin\IEEEauthorrefmark{1}\IEEEauthorrefmark{2},
Liquan Wang\IEEEauthorrefmark{1}\IEEEauthorrefmark{2},
Stavros Tsogkas\IEEEauthorrefmark{1}\IEEEauthorrefmark{3},
Sven Dickinson\IEEEauthorrefmark{1}\IEEEauthorrefmark{2}\IEEEauthorrefmark{3} and
Animesh Garg\IEEEauthorrefmark{1}\IEEEauthorrefmark{2}\IEEEauthorrefmark{4}}
\IEEEauthorblockA{\IEEEauthorrefmark{1}University of Toronto,
\IEEEauthorrefmark{2}Vector Institute,
\IEEEauthorrefmark{3}Samsung AI Center Toronto, \IEEEauthorrefmark{4}Nvidia}
\texttt{\{\href{mailto:dturpin@cs.toronto.edu}{dylanturpin},
\href{mailto:liquan.wang@cs.toronto.edu}{liquan.wang},
\href{mailto:tsogkas@cs.toronto.edu}{tsogkas},
\href{mailto:sven@cs.toronto.edu}{sven}, \href{mailto:garg@cs.toronto.edu}{garg}\}@cs.toronto.edu}}

\maketitle

\IEEEpeerreviewmaketitle

\input{1-abstract.tex}
\input{1-introduction.tex}
\input{2-related_work.tex}
\input{3-problem.tex}
\input{4-method.tex}
\input{5-experiments.tex}
\input{6-conclusion.tex}

\small{
\section*{Acknowledgment}
DT was supported in part the NSERC CREATE in Healthcare Robotics (HeRo) Grant. AG was supported, in part, by the CIFAR AI Chair Grant. The authors would also like to acknowledge the feedback from members of the \href{http://pair.toronto.edu/}{PAIR} research group at UofT.
}

\renewcommand*{\bibfont}{\small}
\bibliographystyle{plainnat}
\bibliography{references}

\input{7-appendices.tex}

\end{document}

%% file: 0-format_and_definitions.tex
\usepackage[pdftex]{graphicx}
\usepackage{epsfig}     % for figures
\pdfoutput=1
\graphicspath{{./figures/}}
\DeclareGraphicsExtensions{.pdf,.jpeg,.png,.jpg}

\usepackage{subcaption}
\usepackage{float}
\usepackage[font={small}]{caption}
\usepackage{lscape}      %Useful for wide tables or figures.
\usepackage{etoolbox} % fix space between abstract header and abstract body

\usepackage[ruled, vlined, linesnumbered]{algorithm2e}

\usepackage{times}
\usepackage{microtype}
\usepackage[utf8]{inputenc} % allow utf-8 input
\usepackage[T1]{fontenc}    % use 8-bit T1 fonts
\usepackage{units}

\usepackage{bm}                          % Make bold, italic math symbols
\usepackage{mathtools}
\usepackage{amsmath, amssymb, amscd}
\usepackage{wasysym} %special symbols
\usepackage{nicefrac}       % compact symbols for 1/2, etc.
\usepackage{mdwmath}
\usepackage{mdwtab}
\usepackage{eqparbox} %equations
\usepackage{array} %equations
\usepackage{nicefrac}

\usepackage{array} %for table entries to be in center of cell
\usepackage{tabularx}
\usepackage{multirow}
\usepackage{multicol}
\usepackage{booktabs}   % Publication quality tables
\usepackage{listings}
\usepackage{nameref}
\usepackage{paralist}
\let\labelindent\relax
\usepackage{enumitem}

\usepackage{xspace}
\usepackage{setspace}

\usepackage{url}  % Hyphenation of URLs.
\usepackage{color} %superceded by xcolor
\usepackage[xcdraw]{xcolor}
\usepackage{hyperref}
\hypersetup{
    unicode=true,
    pagebackref=true,
    breaklinks=true,
    bookmarks=false,
    colorlinks=true,
    pdfborder={0 0 1}
}
\usepackage{comment}
\usepackage{soul} %text highlighting

\newcommand{\norm}[1]{\left \lVert #1 \right \rVert} % norm

\makeatletter
\let\orgdescriptionlabel\descriptionlabel
\renewcommand*{\descriptionlabel}[1]{%
  \let\orglabel\label
  \let\label\@gobble
  \phantomsection
  \edef\@currentlabel{#1}%
  \let\label\orglabel
  \orgdescriptionlabel{#1}%
}
\makeatother

\newcommand{\kps}{K} % keypoints
\newcommand{\kp}[1]{x_{#1}} % keypoint with index i
\newcommand{\gnn}{G} % gnn function name
\newcommand{\gnnparams}{\theta} % gnn function params

%% file: 1-abstract.tex
\begin{abstract}
% \vspace*{-20pt}
Tool use requires reasoning about the fit between an object's affordances and the demands of a task.
Visual affordance learning can benefit from goal-directed interaction experience, but current techniques rely on human labels or expert demonstrations to generate this data.
In this paper, we describe a method that grounds affordances in physical interactions instead, thus removing the need for human labels or expert policies.
We use an efficient sampling-based method to generate successful trajectories that provide contact data, which are then used to reveal affordance representations.
Our framework, \algoName, operates in two phases: first, we discover visual affordances from goal-directed interaction with a set of procedurally generated tools;
second, we train a model to predict new instances of the discovered affordances 
on novel tools in a self-supervised fashion.
In our experiments, we show that \algoName can 
leverage a sparse keypoint representation to predict grasp and interaction points to accommodate multiple tasks, such as hooking, reaching, and hammering. \algoName outperforms baselines on all tasks and matches a human oracle on two of three tasks using novel tools. 
Qualitative results available at:
\href{https://www.pair.toronto.edu/gift-tools-rss21}{www.pair.toronto.edu/gift-tools-rss21}.
\end{abstract}

%% file: 1-introduction.tex
\section{Introduction}
Consider using the hammer in the middle column of Figure~\ref{fig:teaser} to drive a peg into a hole.
This task can be completed in multiple ways;
for instance, one can grip the handle high, for more precise tapping, or low, for greater leverage.
Either face of the head could be used as the point of contact as both provide flat, hard striking surfaces.
Moreover, one can use a hammer not just for this particular task, but also to reach, move, or pry open objects.

This simple thought experiment provides an example of reasoning about what Gibson called \emph{affordances}:
the action possibilities offered by an object~\cite{gibson1977theory,gibson1979ecological}.
Gibson's framework explores the connection between tasks, skills, and tools,
defining affordances only based on whether an action can be completed with a particular tool,
without imposing explicit constraints on the object's appearance, physical, or geometric properties.
Simply put, a hammer can be defined as \emph{any} object that ``affords'' the action of hammering.

This definition motivates the class of \emph{behaviour-grounded} affordance representations.
If affordances are action possibilities, then an ideal affordance representation should be translatable to realizable actions.
This elegantly matches Gibson's definition and allows us to
reality-test our affordances at any time, by simply
executing the corresponding action and observing the result.
In fact, this ``predict, act, and check'' procedure should be sufficient to learn an
affordance representation in a self-supervised manner.
However, the space of possible actions to try is prohibitively large and translating from
an affordance representation to corresponding motions is far from trivial.

Prior work has avoided this roadblock in two ways:
(1) with human supervision~\cite{manuelli2019kpam,allevatolearning}; or
(2) by greatly constraining the space of possible actions~\cite{qin2019keto,fang2018learning,fang2020learning,Mo21Where2Act,mar2017}.
Although labelled data (e.g., keypoint annotations where an object should be grasped and interacted with) remove the need to sample actions,
they can be expensive, time-consuming to collect and may encode irrelevant human biases.
Limiting the action space makes learning from sampled actions tractable, but correspondingly limits the set of
learnable affordances.
Trying a subset of possible actions can only ever reveal a subset of possible affordances.
Instead, we present the first work to tackle this problem without such restrictive simplifications.
By efficiently generating trajectories with a simple sampling-based method
and recovering affordances from contact data, we can achieve representation learning
without human supervision and without constraining the space of possible manipulation actions.

Rather than relying on labels, we learn from goal-directed interaction experience.
To collect experiences, prior self-supervised works design motion generation routines that take an interpretable set of arguments
(e.g., a list of keypoints).
An argument setting is sampled and fed to the motion generation routine
which computes a plan.
The plan is executed and, if the task is successfully completed, the argument setting becomes a positive training example.
This successfully avoids sampling from a large action space,
but comes at the cost of constraining the possible manipulation actions by
``baking in’’ a strategy for completing the task.
For instance, the motion generation routine might specify how to swing a hammer,
given a proper choice of grasp, function and effect points~\cite{qin2019keto}.
The affordance model can then learn to fill in these slots, but can never
discover affordances corresponding to other manipulation strategies.

Instead, we limit our designer intervention to writing a reward function that specifies the task
(and not a specific strategy for completing it).
We generate experiences according to this reward with a simple sampling-based
motion planner in an unconstrained action space.
Our trajectory sampler is free to use tools in unexpected ways, discovering new affordances that
could not have been explored if we were tied to a predefined manipulation strategy.
Furthermore, we ground our affordance representation not on argument settings
but on the contacts between objects.
This makes our method agnostic to the source of trajectories
and avoids the inefficiency of a guess-and-check algorithm.
Instead of testing a predicted affordance by precisely executing the associated action,
we generate a successful action as close as possible to the predicted
affordance and use that to update our model.

Our method for self-supervised discovery of
{\bf G}eneralizable {\bf I}nteraction-aware {\bf F}unctional {\bf T}ool affordances (\textbf{GIFT}) decomposes the affordance representation problem into
i) task-agnostic perception of geometry; and
ii) task-specific affordance detection.
This decomposition allows us to train a single \textit{shared} perception network,
that represents tool geometry as a set of sparse keypoints,
and multiple, smaller, task-specific networks
that detect affordance instances.
The task-agnostic keypoints provide a sufficient representation of a tool's geometry for a variety
of downstream manipulation tasks.
To apply our method to a new task, we need only train a new task-specific network,
without the need to design a new motion generation function or train the perception module from scratch.

We validate the efficacy of our approach with experiments across three manipulation tasks:
hooking, reaching and hammering.
These tasks are basic but learning a generalizable model that can complete them
with unseen tools remains an open problem.
Each task requires using tools in a different way and so experience with each task reveals a different affordance.
Our results show that we match the performance of a human oracle on two out of three tasks
and beat baselines on all three.
We also show, qualitatively, that our method identifies human-like grasp and interaction
keypoints on novel objects that vary by task, as expected.

%% file: 2-related_work.tex
\section{Related Work} \label{sec:related_work}
\subsection{Keypoint learning}
Choosing an appropriate object representation can significantly impact a model's performance in downstream tasks,
such as registration, tracking, and robot manipulation.
A popular choice of object representation in recent years has been 2D and 3D object \emph{keypoints}.
These representations boast several appealing features: they are sparse, efficient, and interpretable,
and they often capture meaningful concepts, such as joints or object parts.
They can also be reliably extracted from data of varying modalities, such as RGBD images or point clouds,
using established CNN backbones~\cite{qi2017pointnet}.
Such modules can be incorporated into a larger pipeline and learn to extract keypoints that
are optimal for the task at hand, often in a \emph{self-supervised} manner, using task-specific training objectives.

KeypointNet~\cite{suwajanakorn2018discovery} learns category-specific keypoints, using 3D pose estimation as the downstream task.
The training objective penalizes inconsistent keypoint predictions
in two different views of the same object,
exploiting the (known) rigid transformation between the two views as a supervisory signal, thus requiring no ground truth keypoint annotations.

6-PACK~\cite{wang20206} employs a similar idea for real-time tracking of novel object instances.
The multi-view consistency loss is computed between keypoints generated in consecutive
video frames, transformed using the ground truth inter-frame motion,
allowing for unsupervised training of the keypoint network.

Transporter~\cite{kulkarni2019unsupervised} is another architecture that discovers keypoints as
geometric object representations in an unsupervised way.
It does so by learning to transform a source video frame into a target frame by
``transporting'' image features at keypoint locations discovered by KeyNet~\cite{jakab2018unsupervised}.

For keypoint prediction, we use an architecture inspired by 6-PACK, but trained using a novel combination of losses to encourage keypoints
that represent tool geometry and serve as possible grasp and interaction points.

\subsection{Affordance learning}
Another relevant line of work aims to learn to perceive the affordances of objects.
Gibson~\cite{gibson1977theory,gibson1979ecological} defined the affordances of an object as
the action possibilities the object offers to an agent in its environment.
This definition suggests that affordances can be recovered by trying actions and observing the results.
In practice, this approach runs into difficulties, since the space of possible actions is prohibitively large.

Given the difficulty of discovering affordance by interaction,
many prior works take a supervised approach.
\cite{myers2015affordance,do2018affordancenet} train on pixel-level affordance labels
to learn to segment images by affordance.
\cite{song2010learning,dang2012semantic,liu2019cage} all learn to predict task or context-appropriate semantic grasps from labelled examples.
Unlike these methods, which all rely on human-provided labels or examples, our method does not require any human supervision.

Our work follows a line of research on behavior-grounded affordance representations,
so-called because their affordance encodings correspond directly to realizable actions.
This elegantly aligns with Gibson's definition of affordance as ``action possibility'', and
has the added benefit that affordance representations can be tested against reality
by executing the associated action.
However, ``behaviour-grounded'' does not equate to ``self-supervised''.
For example, kPAM~\cite{manuelli2019kpam} learns to detect affordance keypoints that are used to
generate motion to complete a hanging task (with motion generated according to constraints defined on the keypoints).
kPAM succeeds in connecting representation to action, but the keypoints are still learned
from human labels.

The work that comes closest to addressing our problem setup is KETO~\cite{qin2019keto}, which learns keypoint affordance representations for tools \emph{without} relying on labels.
KETO extracts features using a PointNet~\cite{qi2017pointnet} backbone, computes keypoint proposals,
and selects three tool keypoints: (grasp, function, effect).
A manipulation action is computed from these and keypoints that result in task success are used
as positive examples for training the keypoint generator.
To get from keypoints to a manipulation action, a final tool pose is computed by solving a quadratic programming problem
parameterized by the keypoints.
The problem is designed to encode an objective of delivering force to a target location
along a target direction, with different targets set for each task.
Specifically, the objective is to align the function point with the target location and align
the vector running from function to effect point with the target direction.

We argue that the design of this problem encodes a strategy that is itself most of a solution to the manipulation task.
For instance, in the hammering task, the quadratic problem encodes the objective of delivering force to the peg
in the direction of the box.
It remains for the system to learn how to select appropriate keypoints as parameters to the problem,
but the outline of a solution is given by this objective.
The learned model is tasked only with filling in the blanks.
Furthermore, keypoint selection is bootstrapped from a heuristic policy that specifies how these points should be chosen.
Results after training do outperform the heuristic, but the framework requires specifying both a manipulation
strategy parameterized by keypoints and a heuristic for choosing those keypoints.
The affordance representation does not emerge only from the constraints of the task.
It may be biased by these prespecified strategies and heuristics.

The \textbf{GIFT} framework addresses this limitation
and allows unbiased affordances to be discovered from free interaction.
We limit our designer intervention to task specification (that is, specifying reward functions
over states, e.g. to maximize peg acceleration)
rather than motion objective specification (that is, objective functions over actions,
as in KETO).
We are not limited to using only rigid transformations~\cite{manuelli2019kpam}, a small set of discrete actions~\cite{mar2017}, or pre-defined motion primitives~\cite{fang2020learning,fang2018learning}.
Instead, we sample trajectories, conditioned on the predicted keypoints, and
score them with a reward function that encourages task completion to refine our original predictions.
The trajectory sampling routine is free to use the tool in unexpected ways,
so we discover affordances without bias from a predefined  motion function or human labels.
In fact, since our keypoints are grounded in the contacts between objects, GIFT is agnostic to the source of trajectories,
as long as we can extract contact data.

Additionally, GIFT exploits the shared representational requirements across tasks by decomposing the affordance discovery problem into inferring:
i) a task-\emph{agnostic} geometry representation; and
ii) a task-\emph{specific} affordance representation.
This allows us to share a single task-agnostic sparse keypoint network across tasks.
Prior supervised works \cite{myers2015affordance,do2018affordancenet} follow a similar decomposition,
but we present the first self-supervised routine
to separate geometry and affordance representations.

%% file: 3-problem.tex
\begin{figure*}[t!]
    \setcounter{figure}{1}
    \centering
    \includegraphics[width=\linewidth]{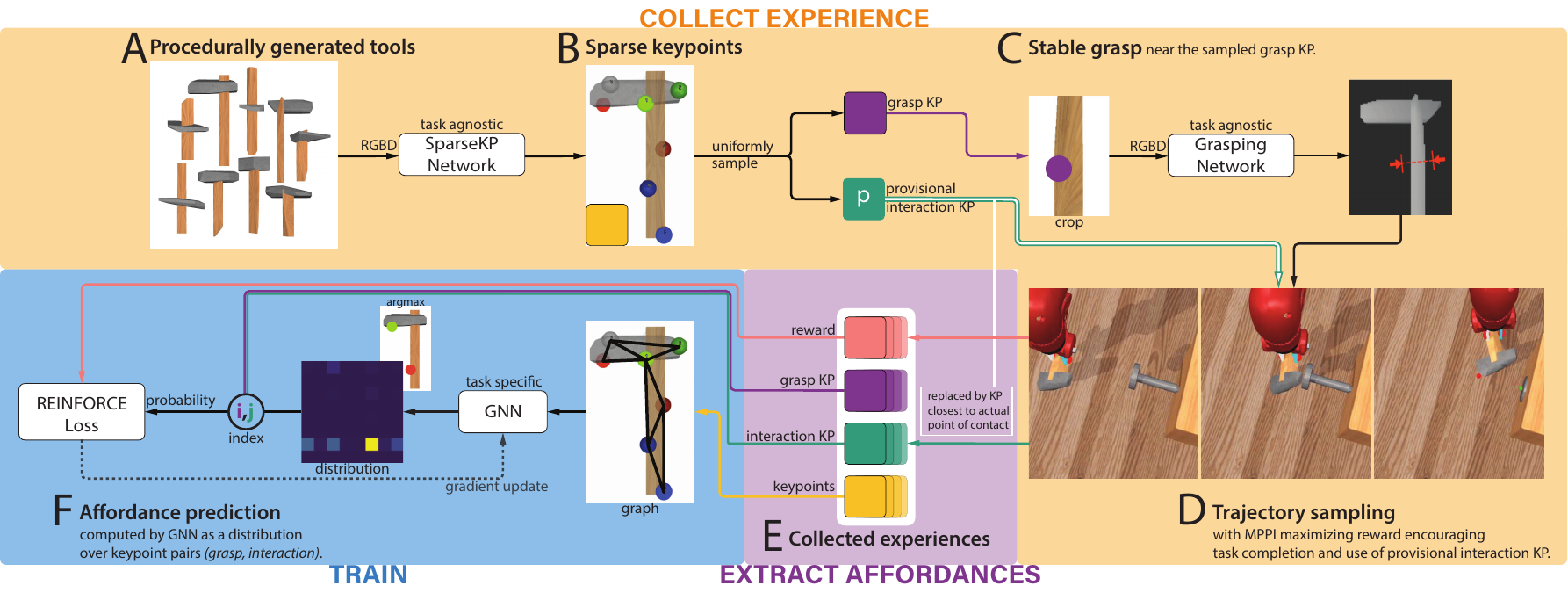}
    \caption{\textbf{\algoName Overview}.
    Our framework learns affordance models for hooking, reaching and hammering by interacting with a set of tools.
    All three models share a task-independent keypoint detector, which takes an RGBD image of a tool and
    predicts a set of keypoints representing that tool's geometry and providing possible choices of grasp and interaction regions.
    The task-conditional portion of each model, which is trained on the outcome of trajectories collected from motion planning,
    selects two keypoints which become our functional tool representation.
    \label{fig:pipeline}
    }
\end{figure*}

\section{Problem Formulation}
We consider the problem of learning affordances without any human labelled data.
We break the problem into two parts: (1) discovering visual affordances through goal-directed interaction with a set
of tools; and (2) learning to predict instances of the discovered affordances from observations of \emph{novel}
tools, with which we have no previous experience.

Since we are not given labelled examples to learn from,  we need a way to generate training data.
For this purpose, we assume access to physics simulations that define a set of manipulation tasks, each with an observable objective,
such as successfully hammering a peg into a block.
We will use these simulators to gather experiences from which we can extract affordance instances.

Each task has a similar structure:
a Sawyer robot arm must grasp a tool lying flat on a table, and use it to manipulate a target object
to complete an objective.
The tasks are divided into families according to their affordance; for example,
all tasks in the ``hammer'' family consist of using a tool to hammer in a peg.
What varies between tasks in a family is the choice of tool.
The challenge is to learn an affordance representation that enables generalization to unseen tools, with novel geometry.

To discover affordances from experience, it is first necessary to build a bank of experiences.
We represent experiences as trajectories, i.e., sequences of simulator states,
with different experiences yielding different affordance instances.

Formally, we define an \emph{affordance instance} as a tuple of two 3D keypoints, $(\kp{grasp},\kp{inter})$,
for grasping the tool and interacting with the target object, respectively.
Previous works like KETO~\cite{qin2019keto} and kPAM~\cite{manuelli2019kpam} have shown that sparse representations are
sufficient for manipulation tasks like the ones we consider.
The two keypoints reflect the two necessary points of contact in a
tool manipulation task: hand-to-tool and tool-to-target.

A trajectory is \emph{compatible} with an affordance instance if during the trajectory
the tool is grasped close enough to $\kp{grasp}$ and contact is made close enough to $\kp{inter}$.
An affordance instance that admits a compatible \textit{and} successful trajectory is called \emph{realizable}.

After collecting experiences and extracting affordance instances from them, one can proceed to
learn the regularities of both grasping and interaction, in terms of keypoints,
that define the affordance across different tool instances,
enabling realizable affordance prediction for novel objects.

%% file: 4-method.tex
\section{Method} \label{sec:method}

\textbf{Overview.}
Figure~\ref{fig:pipeline} provides an overview of our approach, which consists of three stages.
First we collect experiences conditioned on sampled grasp and interaction keypoints.
Our SparseKP Network detects a set of keypoints from an RGB-D observation of a tool.
These represent task geometry and provide labelled candidate areas on which to grasp the tool or interact with the world.
From these we sample a grasp and interaction point.
We execute a stable grasp near the grasp point and sample the rest of the trajectory with MPPI
using a reward that encodes task success and encourages use of the sampled interaction point.

Next we extract affordances from the sampled trajectories which become training examples for the affordance model.
Finally, we use the extracted training examples to train an affordance model using a REINFORCE loss.
The affordance model is a graph neural network (GNN) whose input is a graph defined over the keypoints $\kps$.
The model computes a probability distribution over combinations of keypoint pairs,
reflecting their suitability as grasp and interaction points, in the context of the entire graph.
The weights of the model are optimized so that high probability is assigned to keypoint pairs
that are compatible with a trajectory that allows task completion, thereby providing the desired affordance.
In this way, the model implicitly learns how to map tool geometry to an affordance, and evaluate
that affordance on new tool geometries.

\subsection {Keypoint-based Sparse Object Representations}
\label{sec:method:keypoints}

\textbf{Keypoint detection pipeline.}
Given an RGB-D observation of a tool, we sample a point cloud from the depth channel.
We are not given a tool segmentation, but we do mask out the most common value when sampling the point cloud from the depth image, since we assume this value corresponds to the table.
We use a PointNet~\cite{qi2017pointnet} backbone to encode local geometric features of the point cloud,
and a ResNet~\cite{he2016deep} to compute appearance features from the RGB channels.
We concatenate the geometric features of each sampled point and the appearance features of its corresponding pixel in the image,
and pass the combined result to a final module that predicts $M$ keypoints $\kps = \{\kp{1}, \ldots, \kp{M}\}$.

Our keypoint network draws inspiration from 6-PACK~\cite{wang20206}.
Whereas \cite{wang20206} aimed to detect keypoints for pose estimation, our goal is to discover keypoints that capture tool geometry and provide possible grasp and interaction points.
Specifically, we want keypoints that are spread across the tool and lie near geometric features, such as edges and corners.
To this end we choose a different combination of losses, namely a quadric loss~\cite{agarwal2019learning} and a coverage loss.

\begin{figure*}
    \centering
    \begin{subfigure}[b]{0.325\textwidth}
        \includegraphics[width=\textwidth]{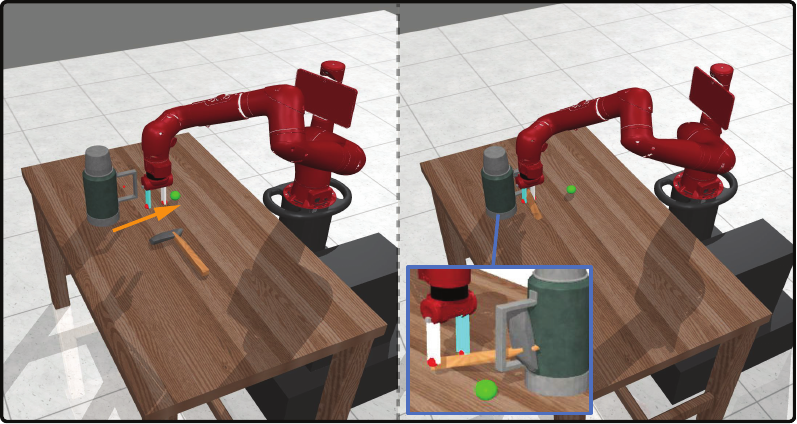}
        \caption{Hooking}
        \label{fig:hooking}
    \end{subfigure}
    \begin{subfigure}[b]{0.325\textwidth}
        \includegraphics[width=\textwidth]{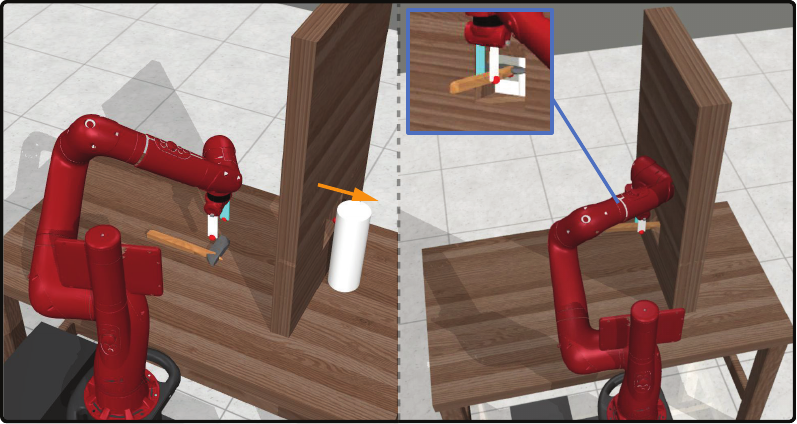}
        \caption{Reaching}
        \label{fig:reaching}
    \end{subfigure}
    \begin{subfigure}[b]{0.325\textwidth}
        \includegraphics[width=\textwidth]{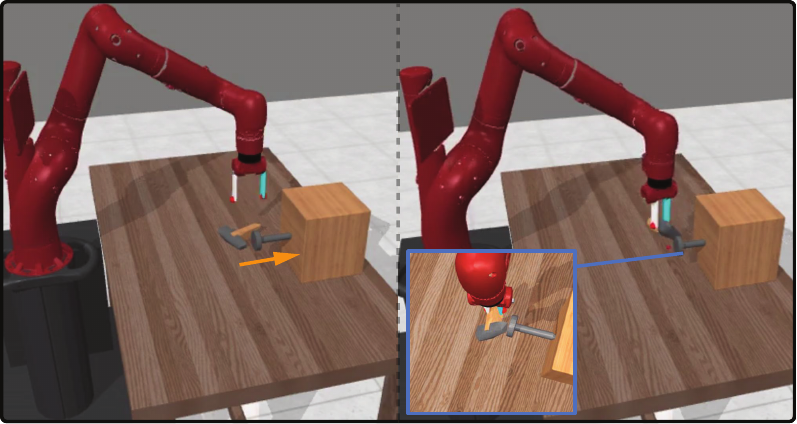}
        \caption{Hammering}
        \label{fig:hammering}
    \end{subfigure}
    \caption{\textbf{Interaction based manipulation tasks.} \algoName learns emergent interaction-aware tool affordances on three tasks: hooking, constrained-reaching, and hammering. The same set of procedurally generated tools is used across tasks.}
    \label{fig:tasks}
\end{figure*}

\textbf{Coverage loss.}
The coverage loss encourages keypoints to be spread over the tool surface,
avoiding solutions where keypoints are concentrated in a very small neighborhood.
Initially we experimented with a simple separation loss, that encouraged the pairwise
Euclidean distance between keypoints to be above some threshold.
We found this loss competed with the quadric loss, making training difficult.
The separation loss pushed keypoints apart, possibly off the tool, and the quadric loss
pulled them towards the tool surface.

We introduce a coverage loss that does not compete with the quadric loss.
Each keypoint $\kp{} \in K$ induces a distribution over mesh vertices $v \in V$,
with probability inversely proportionate to the Euclidean distance to the keypoint:
\begin{equation}
    P_x(v) = \frac{\exp{(-\norm{x - v}_2})}{\sum_j \exp{(-\norm{x - v_j}_2)}}.
    \label{eq:distribution}
\end{equation}
We want the average of these distributions to be uniform, so that each vertex has a similar
average distance to keypoints, giving good coverage of the object.
We also want to take the tool's geometry into account when computing the distance between
distributions.
Specifically, we want to minimize the Earth Mover's Distance (EMD)~\cite{rubner2000earth}
between the average of the induced
distributions and the uniform distribution, with geodesic distance as our ground distance.
Letting $D_K$ be the average of these distributions, given by $P_K(v) = \frac{\sum_i
P_{x_i}(v)}{\norm{K}}$,
and $D_U$ be the uniform distribution, given by $P_U(v) = \frac{1}{\norm{V}}$,
we minimize the Sinkhorn
distance~\cite{cuturi2013sinkhorn}, which is a differentiable approximation to the EMD:
\begin{align}
   \mathcal{L}_{coverage} = \textrm{Sinkhorn}(D_K, D_U | G),
   \label{eq:coverage}
\end{align}
where $G$ is a precomputed  matrix of pairwise geodesic distances.
Our approach is inspired by \cite{solomon2014earth} which uses a similar technique
to extend the geodesic distance along a mesh to points off the mesh surface.

\textbf{Quadric loss.}
The quadric loss encourages keypoints to be near edges and corners of the tool's surface,
since we expect keypoints near edges and corners to better capture the geometric features of the tool.
It is based on the quadric error, which measures the distance between a point and a plane.
Given a tool mesh with vertices $V$ and faces $F$, and a keypoint $x$, we
define its nearest vertex $v_{x} = \min_{v \in V} \norm{x - v}_2$.
Let $\{p_j\}$ be the planes of faces incident to $v_x$, with $p_j = [a, b, c, d]^T$.
The squared distance to each plane can be calculated as a product between the keypoint and the plane's quadric matrix $q_j = p_jp_j^T$, as $x^T q_j x$.
The quadric loss for a keypoint $x \in K$ is the sum of this distance over all planes of incident faces $\sum_j x^T q_j x = x^T Q_x x$, with $Q_x = \sum_j q_j$.
Then the full quadric loss is the sum of this quantity over keypoints:
\begin{align}
    \mathcal{L}_{quadric} = \sum_i \kp{i}^T Q_{\kp{i}} \kp{i}.
    \label{eq:quadric}
\end{align}

\subsection{Grasp planning}
Given the predicted keypoints, $K$, we uniformly sample a grasp keypoint $\kp{grasp}$, near which we will grasp.
To plan a grasp, we use a pretrained Fully Convolutional Grasp Quality CNN (FC-GQ-CNN).
This is an extension of DexNet 4.0~\cite{satish2019policy} that does not require an initial analytic grasp sampling step.
Instead of scoring a sampled grasp, FC-GQ-CNN densely evaluates possible grasps (positions and z-axis rotations) with a fully convolutional filter.\footnote{For simplicity, we refer to this model as DexNet.}
Since we want DexNet to sample grasps near the selected grasping keypoint,
we take a crop of the RGBD tool observation, centered around $\kp{grasp}$ and pass the crop to DexNet.
DexNet generates 50 candidate grasps from which we select the closest to the grasp keypoint
that is likely to succeed (i.e., has a DexNet Q-value $>0.9$).

We execute the grasp with a simple procedure that moves the end effector over
the center of the grasp, rotates to the match grasp z-axis angle,
lowers to the grasp depth, closes the gripper, and lifts.
Before motion planning takes over, the routine moves the object such that the chosen striking point is in front of the target.
Motion planning proceeds only if the tool is not dropped, since the task can
only be completed if grasping was successful.

\subsection{Motion planning}

\textbf{Trajectory sampling.}
Given the predicted keypoints, $K$, we uniformly sample a provisional interaction keypoint $\hat{\kp{}}_{inter}$.
We use model-predictive control with a sampling-based motion planner to generate the rest of the manipulation motion.
The objective of the planner is to maximize a task-specific reward function.
The planner proceeds iteratively by  planning over a horizon of $H$ timesteps, executing the first step of the plan and then replanning.

Planning begins with a base trajectory of length $H$ with all actions set to $0$.
The action space is the change in $(x, y, z)$ position and $z$-axis rotation of the end effector.
We then generate a set of $M$ perturbed trajectories, by adding Gaussian noise to the base trajectory.
The perturbed trajectories are executed in the simulator and scored using the reward function.
We use a reward-weighted average of these trajectories as the plan and execute its first step
before the whole process repeats.
Specifically we use the MPPI~\cite{williams2017information} planner implemented in LyceumAI~\cite{summers2020lyceum}.

\textbf{Reward formulation.}
The reward function of each task is composed of two terms:
the first term encourages task completion;
the second term encourages the minimization of the distance between the provisional interaction point
and the target object.
The full reward function  for task $T$ can be written as
\begin{equation}
    R_T = C_T - \tanh{(\norm{\hat{\kp{}}_{inter} - x_{target}}_2)},
    \label{eq:reward}
\end{equation}
where $C_T$ is the task completion reward.
The magnitude of the second term is typically low, allowing the planner to explore potentially better
interaction points, i.e., ones that result in higher task-completion reward, near $\hat{\kp{}}_{inter}$.
For each task family considered, we define a task-specific performance reward that is shared across all instances (i.e., tools) in that family.
The rest of the framework remains unchanged between tasks.

Choosing the right reward function impacts performance.
For instance, in our hammering experiments, we found that labellers consistently select a low grasp on the hammer, suggesting they want to maximize leverage.
A reward function that only takes into account the peg position
specifies a task that can be solved by pushing.
Unlike swinging, pushing does not exploit leverage,
so does not constrain the choice of grasp.
As long as pushing solves the task, we can not recover the leverage-maximizing grasp regularity from experience.

To prevent pushing-based strategies, we use a reward based on the instantaneous acceleration of the peg into the block at the moment the tool first makes contact with the peg:

\begin{equation}
    C_{hammer} = \begin{cases}
        (\overrightarrow{a_{peg}}^{goal})^2 &\text{if first contact}\\
        0 & \text{otherwise}
        \end{cases}
\end{equation}
Swinging the hammer accumulates kinetic energy in the head of the tool, which is quickly released when contact is made with the peg, maximizing instantaneous acceleration.

Our reward specification is a much weaker signal
than labelling affordances and/or keypoints and it is generic across the task family.
In tandem with the motion planner, our rewards provide a rich loss function for our
perception network, which is trained without expert knowledge or feature engineering.
For instance, we do not specify that the strike point for hammering should be on the hard
heavy head of the hammer, but we do specify that maximizing peg acceleration is desirable
for hammering, however it can be achieved.
We find such ``designer intervention'' both appropriate and practical, since even the most
advanced RL methods struggle solving these tasks when using a
purely binary sparse reward, especially with a large variety of object instances.

Once we have computed the trajectory, we discard the provisional interaction point $\hat{\kp{}}_{inter}$,
and replace it with the keypoint closest to the first point of  contact between the tool and the target in the sampled trajectory.
The result of this procedure is a tuple $(K, \kp{grasp}, \kp{inter}, R)$, which is added to our training set.

\subsection{Training the affordance model}\label{sec:method:affmodel}
We use a graph neural network (GNN), $\gnn(\theta) = \gnn_\theta$, to predict affordance instances.
$\gnn_\theta$ consists of three generalized graph convolution layers~\cite{li2020deepergcn},
each followed by a SAGPool layer~\cite{lee2019self}. We use graph convolutions to propagate information through the graph and pooling layers to incrementally aggregate information into a global encoding.
Starting from our task-agnostic keypoints $K = \{\kp{0}, \ldots, \kp{M}\}$, we build a graph $(K,E)$,
with edges connecting each node to its three nearest neighbours in Euclidean space.
The penultimate layer of the GNN computes local feature encodings at each node,
and a single global encoding for the entire graph.
We also build pairwise features by appending two local node encodings to the global encoding,
and passing the result through a linear layer which scores that particular choice of nodes as
grasp and interaction keypoints.

At training time, we pass $(K, E)$ as input to our affordance model, $\gnn_\gnnparams$,
which computes a probability distribution over pairs of keypoint indices $D = \gnn_\theta(K,E)$.
$D_{ij}$ is the probability of selecting $\kp{i}$ as grasp keypoint and $\kp{j}$ as interaction keypoint.
We aim to find model weights that maximize the expected reward of trajectories matching grasp and interaction keypoints
sampled under this output distribution $D$.
Our optimization objective is the REINFORCE loss~\cite{williams1992simple},
\begin{equation}
    \mathcal{L}_{R} = \sum_{(\kp{g}, \kp{intr}, \kps, R)} -P(\kp{g}, \kp{inter} | D = \gnn_\gnnparams(\kps))R.
    \label{eq:reinforce}
\end{equation}
Implementation details of the method are described in the appendices. 

%% file: 5-experiments.tex
\begin{figure}
\centering
\includegraphics[width=.8\linewidth]{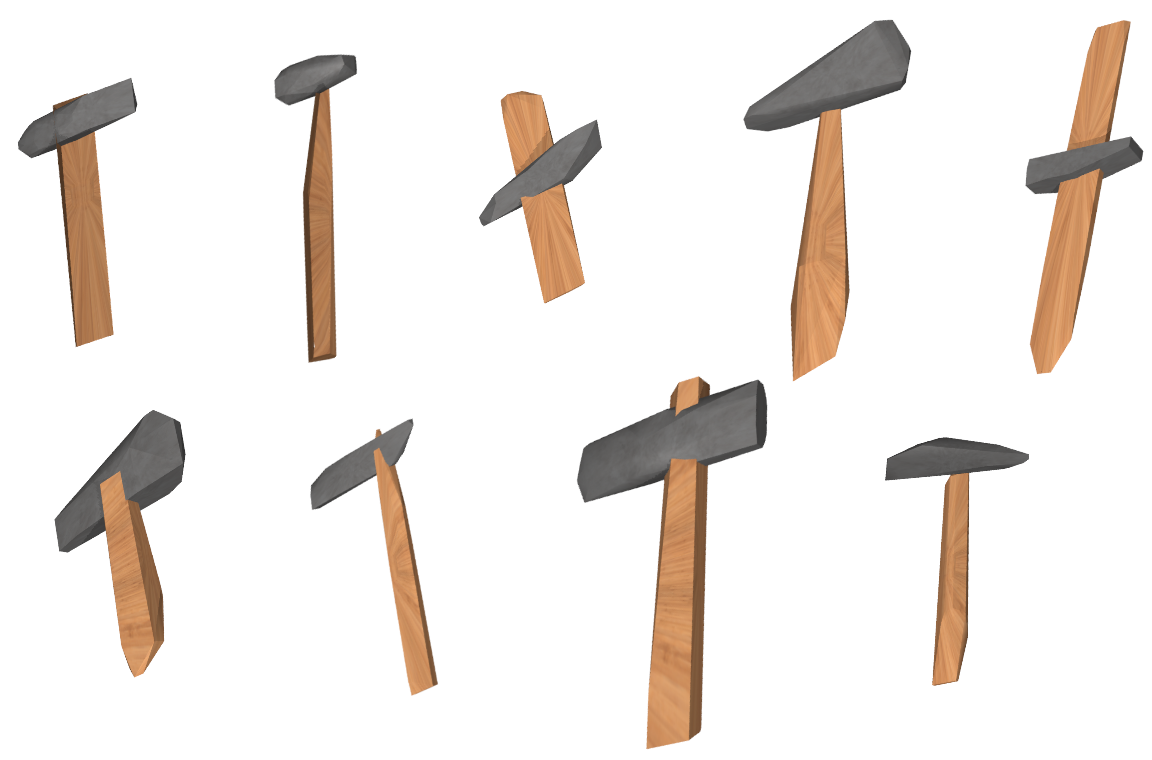}
\caption{\textbf{Procedural Tool Shapes}. A sampling of tools procedurally generated into \texttt{T}, \texttt{X} or \texttt{L} configurations.
We use 10K such shapes to pre-train SparseKP tool representations and another 300 objects to train the task-specific affordance models.
}
 \label{fig:tool-examples}
\end{figure}

\section{Experiments} \label{sec:experiments}

\input{5-expt-table}

We perform quantitative evaluation of \algoName to answer the following questions:
\begin{enumerate}[wide,labelwidth=!,labelindent=0pt]
\item Can \algoName learn task conditional affordances from functional interactions for different tasks?
\item How do learned \algoName representations compare with affordances labelled by a human oracle (qualitatively and in task performance)?
\item How sensitive are the predictions to material properties?
\end{enumerate}

\subsection{Experimental Setup}
We consider three families of manipulation tasks: \texttt{hooking}, \texttt{reaching}, and \texttt{hammering}, as shown in Figure~\ref{fig:tasks}.
These are basic tasks and it is not difficult to train a model to succeed on a single task instance (completing a set task with a set tool).
The challenge is to learn an affordance model that can successfully generalize \emph{across an entire task family} (performing a familiar task with new tools).
All three tasks follow a similar structure:
first the tool must be grasped, lifted, and aligned with the target;
then the tool must contact the target object, and manipulate it towards a goal condition. In all tasks, correct use of the tool enables a mechanical advantage.
For instance, in the hammering task, the goal is to drive a peg into a box.
The task reward encourages maximizing peg acceleration into the block, which requires generating leverage by grasping the tool low on the handle, swinging to accumulate kinetic energy and striking with the head to rapidly transfer that energy into the peg.
In the hooking task, the goal is to move a mug towards a target position by hooking its handle.
To avoid pushing-based solutions, which do not much constrain the choice of grasp and interaction points, we add a penalty for touching the mug anywhere besides the handle.
In the reaching task, the goal is to push an object on the other side of a wall by reaching through a small hole in the wall, which can only be achieved by choosing a thin, long part of the tool to make contact.
We design these interactive tasks in the MuJoCo physics simulator~\cite{todorov2012mujoco} with a Julia interface using Lyceum~\cite{summers2020lyceum}.
We use $M=8$ task-agnostic keypoints.

\textbf{Procedural generation of tools.} \label{sec:experiments:dataset}
\algoName is trained in simulation with a large set of 3D models. However, existing 3D model datasets do not contain enough objects suitable for tool manipulation while exhibiting rich variations in terms of their geometric and physical properties. We follow the procedural tool generation routine from~\cite{fang2020learning}.
To generate varied, complex shapes, we take the convex decomposition of meshes from~\cite{bohg2013data} and merge randomized pairs of convex shapes into a \texttt{T}, \texttt{X} or \texttt{L} configuration (see Figure~\ref{fig:tool-examples}).
We use a dataset of 10,000 objects to train the keypoint generator model.
In the task-conditional stage, 300 objects are used for experience collection.
Finally at test-time, a non-overlapping set of 100 unseen objects are used for evaluation.

\textbf{Baselines.} \label{sec:experiments:baselines}
We compare our \algoName affordance model against four baselines:
\begin{enumerate}[wide,labelwidth=!,labelindent=0pt]

\item \emph{Simple.} A naive baseline is to choose a grasp by sampling antipodal points on the object and  uniformly sampling a keypoint as the interaction point.
An antipodal pair of points have (approximately) opposing normals, hence this yields a reasonable grasping heuristic.
The interaction point is chosen randomly from the sparse keypoint set output by our SparseKP network,
which are plausible candidates as they lie near the object surface.

\item \emph{Grasp Optimized} uses a pre-trained grasping model (DexNet~\cite{mahler2017dex}) to select the globally optimal stable grasp. It chooses the interaction point uniformly, as in the Simple baseline.

\item \emph{Leverage Heuristic.} We improve on the Grasp Optimized baseline by choosing the interaction point to be the farthest keypoint from the grasp point. This is often nearly optimal, especially if the grasp is chosen correctly.

\item \emph{Human Oracle.} We also have a manually labelled dataset with labels for task-conditioned grasp and interaction points. This data is gathered by two labelers with 100 objects for each of the three tasks, establishing an oracle-like performance baseline, useful to evaluate \algoName representations qualitatively.
\end{enumerate}

Unfortunately, it is not possible to directly compare to prior work quantitatively.
For instance, the code for KETO~\cite{qin2019keto} is not publicly available and kPAM~\cite{manuelli2019kpam} uses a substantially different experimental setup, since it requires human labels.
Instead we use \emph{strong} baselines that are as effective or even outperform the above methods.
For example, we argue that kPAM will rarely exceed the performance of a human labeller, such as our human oracle;
and the best performing baselines (e.g., Leverage Heuristic for hammering)
will perform similarly or even better than KETO, since they hardcode expert knowledge.
Picking a stable grasp and a far away strike point is exactly what we expect our method to learn for hammering, so
this baseline provides a meaningful target level of performance.

\begin{figure}
\centering
\includegraphics[width=\linewidth]{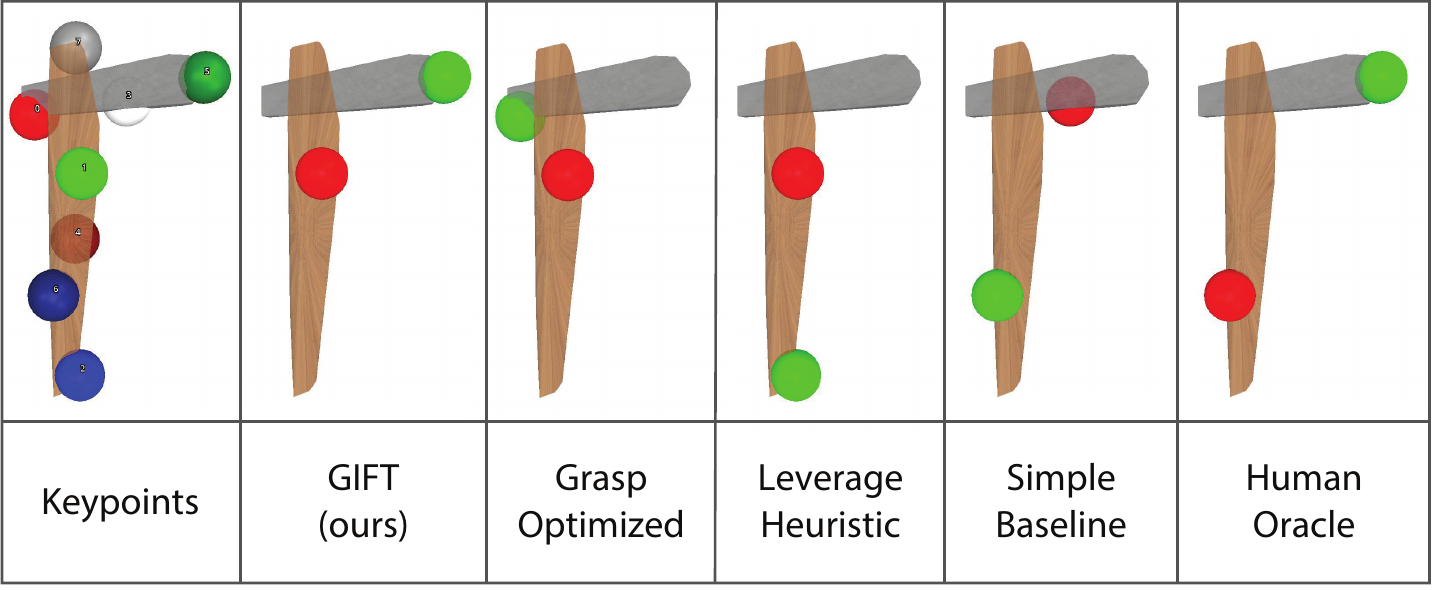}
 \captionof{figure}{\textbf{Baseline Comparison.} Comparison with baselines and a human oracle on the hammering task.
 Whereas the baselines can not consistently identify plausible pairs of grasp and interaction keypoints,
 our method can identify stable grasp points and viable interaction keypoints that better align with human choices.
}\label{fig:qualitative}
\end{figure}

\textbf{Evaluation at test-time.} \label{sec:experiments:testing}
When generating grasps and trajectories for training, the choice of grasp and interaction keypoints acts as
a suggestion -- a prior that suggests ``grasping near here'' and ``making contact near here'', rather than a hard constraint.
In contrast, at test time, we are less flexible.
We want to check whether grasping at the selected grasp point and making contact with the selected interaction keypoint will succeed.
To enforce this, we tweak the reward computation.
If any keypoint is closer to the first point of contact between the tool and target object than $\kp{inter}$,
then the task-specific reward term, $C_T$ in equation~\eqref{eq:reward}, is set to $0$.
This ensures that the trajectory is testing the predicted grasp and interaction points specifically,
rather than treating them as provisional, as during training.

\begin{figure}[t!]
\centering
\includegraphics[width=\linewidth]{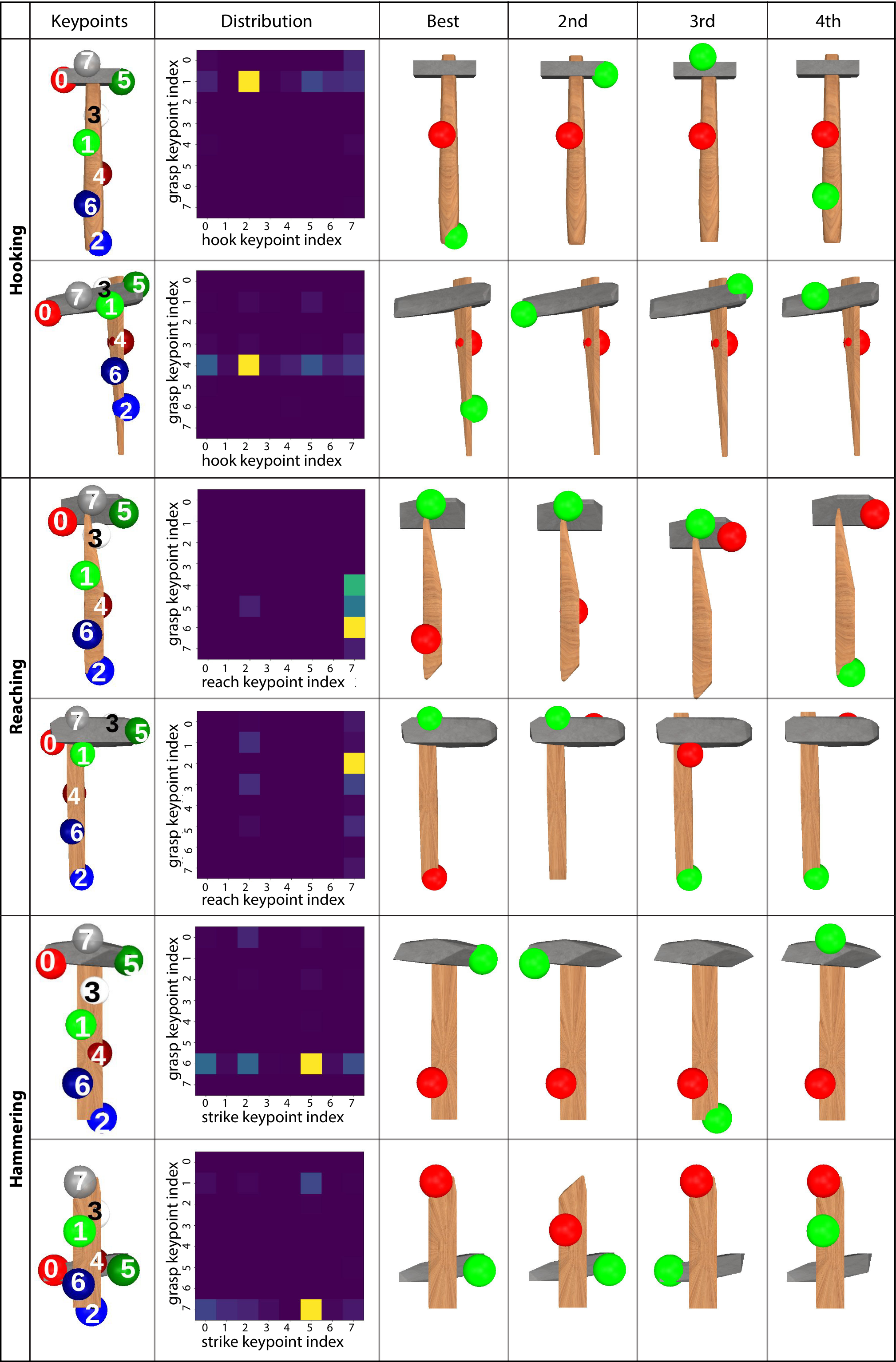}
 \captionof{figure}{\textbf{Results Visualization}. We visualize predicted grasp and interaction keypoints for all three tasks on generated tools. The first column shows predicted task-agnostic keypoints.
 Our affordance model predicts a distribution over pairs of grasp and interaction keypoints selected from these, which is shown in the second column.
 Further columns show the four most likely task-specific keypoint pairs.
 In the top-4 visualizations, the grasping keypoint is shown in red and the interaction keypoint is shown in green.}\label{fig:results}
\end{figure}

\textbf{Metrics.} We evaluate all methods on the following metrics with 4000 episodes each:
\begin{enumerate}[wide,labelwidth=!,labelindent=0pt]
\item \emph{Task Success.} Rate of completion of the overall task.
\item \emph{Mean Reward and Normalized reward.} Reward represents the mean reward over all episodes in the evaluation. A more informative metric is the normalized mean reward with respect to the \textit{Human Oracle}.
\item \emph{Grasp-Conditioned (GC) Success.} Many episodes may fail due to an incorrect choice of grasp, which is often due to inaccuracies of the physics simulator. Hence, we evaluate task success in the subset of cases where the predicted grasp succeeds. This evaluates the affordance model learned through interaction more closely, since failures are not attributed to interaction rather than grasping.
\item \emph{GC-Reward and Normalized Reward.} Mean and normalized reward conditioned on successful grasping.
\end{enumerate}

\subsection{Results}

\noindent \textbf{E1. Learning affordances from functional interactions.} We present the results of self-supervised learning of affordance representations in Table~\ref{tab:my-table}. These results illustrate that \algoName can learn useful affordance mappings that succeed in task completion and generalize to new objects across all three tasks: \texttt{Hooking}, \texttt{Reaching} and \texttt{Hammering}.

Figure~\ref{fig:results} shows affordance instances detected by our method.
For each of our three manipulation tasks, our affordance model selects
some reasonable pairs of grasp and interaction points among their top 4 choices
and often identifies multiple reasonable ways of using a tool (e.g., either side of the head of the hammer in the second-from-bottom
row of Figure~\ref{fig:results} is good for striking).
The model is not perfect and sometimes selects implausible pairs
(as in the 4th best selection of the last row of Figure~\ref{fig:results}),
but usually at least some reasonable choices are present in the top 4.

\noindent \textbf{E2. Comparison of \algoName with human affordance.}
\algoName reprepresentations perform better than carefully optimized baselines
and often reach task performance similar to that of a Human Oracle,
particularly when considering performance conditioned on grasp success.

Figure~\ref{fig:qualitative} shows a qualitative comparison of affordance maps across all baselines and \algoName for the \texttt{Hammering} task.
The Grasp Optimized baseline is able to select stable grasp keypoints,
but cannot distinguish between good and bad interaction keypoints.
\algoName is able to select stable grasps and plausible interaction points that are topologically similar to
those of the human oracle.

\begin{figure}[t!]
\centering
\includegraphics[width=\linewidth]{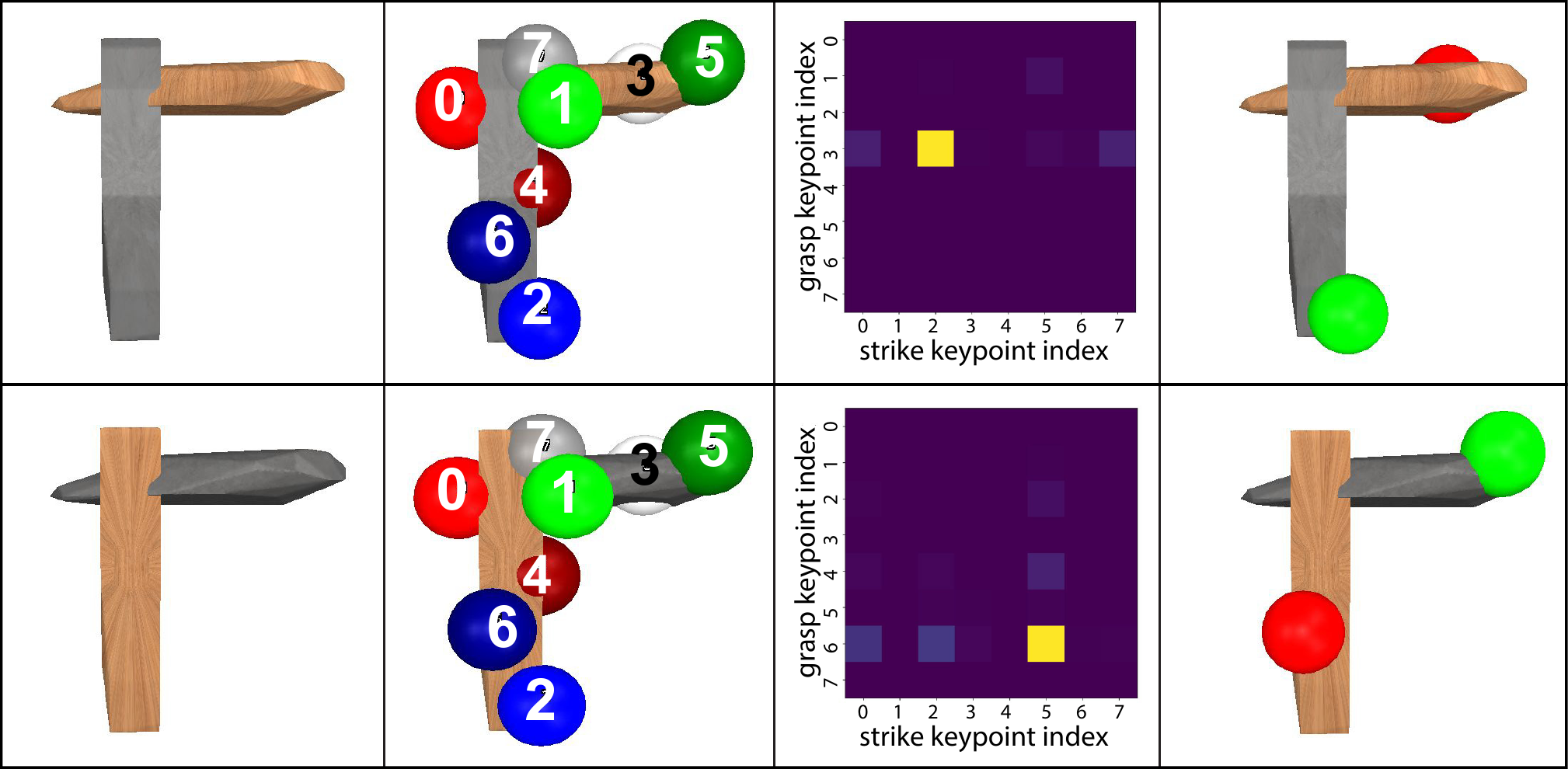}
 \captionof{figure}{\textbf{Comparison with inverted materials}. We test the sensitivity of our framework to material properties, by flipping head and handle materials and retraining our affordance model. For inverse-material hammers, the results show the selection of functional keypoints correctly flips along with the materials. }\label{fig:materials}
\end{figure}

\noindent \textbf{E3. Sensitivity of \algoName to material properties.}
Intuitively, we expect that affordance maps, particularly for tools, depend on material properties and distributions.
For instance, for hammering, a hard dense metal head provides a better strike surface and more moment than a softer, lighter wooden handle.
To evaluate if \algoName will discover such semantics, we retrain with the same tool set, but with materials switched between the head and handle.
We find that the choice of grasp and interaction keypoints also flips to be consistent with the new materials (see Figure~\ref{fig:materials}).

Since material properties are not explicitly encoded, it is possible that the GNN is overfitting to the SparseKP network output, by learning to associate material properties with regularities in keypoint positions. A simple extension of our method with additional property encodings for each keypoint would solve this.

%% file: 5-expt-table.tex
% \usepackage{booktabs}
% \usepackage{graphicx}
\begin{table*}[t!]
\centering
\resizebox{\textwidth}{!}{%
\begin{tabular}{@{}lllrrrrrrr@{}}
\toprule
\multicolumn{1}{l|}{} &  & \multicolumn{1}{c|}{} & \multicolumn{3}{c|}{\textbf{Overall Performance}} & \multicolumn{4}{c}{\textbf{Performance (conditioned on Grasp Success)}} \\ \cmidrule(l){4-10} 
\multicolumn{1}{l|}{\multirow{-2}{*}{}} & \multirow{-2}{*}{\textbf{Grasping Method}} & \multicolumn{1}{c|}{\multirow{-2}{*}{\textbf{Interaction Method}}} & \multicolumn{1}{c|}{\textbf{\begin{tabular}[c]{@{}c@{}}Task \\ Success (\%)\end{tabular}}} & \multicolumn{1}{c|}{\textbf{\begin{tabular}[c]{@{}c@{}}Absolute \\ Reward\end{tabular}}} & \multicolumn{1}{c|}{\textbf{\begin{tabular}[c]{@{}c@{}}Normalized \\ Reward\end{tabular}}} & \multicolumn{1}{c|}{\textbf{\begin{tabular}[c]{@{}c@{}}Grasp \\ Success (\%)\end{tabular}}} & \multicolumn{1}{c|}{\textbf{\begin{tabular}[c]{@{}c@{}}Task \\ Success (\%)\end{tabular}}} & \multicolumn{1}{c|}{\textbf{\begin{tabular}[c]{@{}c@{}}Absolute \\ Reward\end{tabular}}} & \multicolumn{1}{c}{\textbf{\begin{tabular}[c]{@{}c@{}}Normalized \\ Reward\end{tabular}}} \\ \midrule
\multicolumn{10}{l}{\textbf{Hooking}} \\ \midrule
\multicolumn{1}{l|}{Supervised oracle} & Human & \multicolumn{1}{l|}{Human} & 57.9\% & 969.30 & \multicolumn{1}{r|}{1.000} & 74.8\% & 77.3\% & 1,295.42 & 1.000 \\
\multicolumn{1}{l|}{Simple baseline} & Antipodal & \multicolumn{1}{l|}{Random (sparseKP)} & 31.8\% & 534.40 & \multicolumn{1}{r|}{0.551} & 54.4\% & 58.3\% & 982.40 & 0.758 \\
\multicolumn{1}{l|}{Grasp Optimized baseline} & Stability (DexNet) & \multicolumn{1}{l|}{Random (sparseKP)} & 40.1\% & 730.20 & \multicolumn{1}{r|}{0.753} & 69.9\% & 58.6\% & 1,045.20 & 0.807 \\
\multicolumn{1}{l|}{Leverage Heuristic baseline} & Stability (DexNet) & \multicolumn{1}{l|}{Farthest (sparseKP)} & 50.6\% & 887.90 & \multicolumn{1}{r|}{0.916} & 70.1\% & 71.7\% & 1,256.90 & 0.970 \\
\multicolumn{1}{l|}{\algoName (Ours)} & \algoName & \multicolumn{1}{l|}{\algoName} & \textbf{57.9\%} & \textbf{964.50} & \multicolumn{1}{r|}{\textbf{0.995}} & 75.2\% & \textbf{77.0\%} & \textbf{1,282.90} & \textbf{0.990} \\ \midrule
\multicolumn{10}{l}{\textbf{Reaching}} \\ \midrule
\multicolumn{1}{l|}{Supervised oracle} & Human & \multicolumn{1}{l|}{Human} & 60.0\% & 522.31 & \multicolumn{1}{r|}{1.000} & 67.6\% & 89.1\% & 754.00 & 1.000 \\
\multicolumn{1}{l|}{Simple baseline} & Antipodal & \multicolumn{1}{l|}{Random (sparseKP)} & 45.6\% & 312.56 & \multicolumn{1}{r|}{0.598} & 56.6\% & 80.6\% & 632.20 & 0.838 \\
\multicolumn{1}{l|}{Grasp Optimized baseline} & Stability (DexNet) & \multicolumn{1}{l|}{Random (sparseKP)} & 53.1\% & 433.23 & \multicolumn{1}{r|}{0.829} & 64.1\% & 82.9\% & 701.86 & 0.931 \\
\multicolumn{1}{l|}{Leverage Heuristic baseline} & Stability (DexNet) & \multicolumn{1}{l|}{Farthest (sparseKP)} & 54.9\% & 498.12 & \multicolumn{1}{r|}{0.954} & 64.9\% & 84.4\% & 732.30 & 0.971 \\
\multicolumn{1}{l|}{\algoName (Ours)} & \algoName & \multicolumn{1}{l|}{\algoName} & \textbf{60.0\%} & \textbf{512.20} & \multicolumn{1}{r|}{\textbf{0.981}} & 68.3\% & \textbf{87.6\%} & \textbf{734.21} & \textbf{0.974} \\ \midrule
\multicolumn{10}{l}{\textbf{Hammering}} \\ \midrule
\multicolumn{1}{l|}{Supervised oracle} & Human & \multicolumn{1}{l|}{Human} & 75.3\% & 15.67E+6 & \multicolumn{1}{r|}{1.000} & 79.8\% & 94.4\% & 19.69E+6 & 1.000 \\
\multicolumn{1}{l|}{Simple baseline} & Antipodal & \multicolumn{1}{l|}{Random (sparseKP)} & 50.6\% & 6.80E+6 & \multicolumn{1}{r|}{0.434} & 59.6\% & 84.9\% & 11.56E+6 & 0.587 \\
\multicolumn{1}{l|}{Grasp Optimized baseline} & Stability (DexNet) & \multicolumn{1}{l|}{Random (sparseKP)} & 58.9\% & 7.51E+6 & \multicolumn{1}{r|}{0.479} & 68.1\% & 86.5\% & 11.10E+6 & 0.564 \\
\multicolumn{1}{l|}{Leverage Heuristic baseline} & Stability (DexNet) & \multicolumn{1}{l|}{Farthest (sparseKP)} & \textbf{66.1\%} & 10.92E+6 & \multicolumn{1}{r|}{0.697} & 70.9\% & 93.2\% & 15.51E+6 & 0.788 \\
\multicolumn{1}{l|}{\algoName (Ours)} & \algoName & \multicolumn{1}{l|}{\algoName} & 64.6\% & \textbf{11.40E+6} & \multicolumn{1}{r|}{\textbf{0.728}} & 68.9\% & \textbf{93.7\%} & \textbf{16.65E+6} & \textbf{0.846} \\ \bottomrule
\end{tabular}%
}
\caption{We compare the performance metrics: Task Success ($\%$), Absolute Reward and Normalized Reward in three simulated tasks. Since  \algoName is focused on task-completion, not grasp stability, we report performance both overall and conditional on grasp success. The latter is indicative of learning performance since \algoName builds on a pre-trained grasping model. \algoName matches human oracle performance on both hooking and reaching and beats baseline performance on all three tasks.
In all tasks, in cases of successful grasping, learning-based \algoName outperforms all baselines including task-specific hand-crafted heuristics and achieves success rates close to a human oracle.
}
\label{tab:my-table}
\end{table*}

%% file: 6-conclusion.tex
\section{Conclusions}
We have presented a method for the self-supervised discovery of functional tool affordances
that generalize to novel tools.
The ability to use vision to transfer affordances discovered from experience
to observations of new objects is the core of our contribution.
We achieve this without relying on human labels
or a simplified action space.
Instead we collect experience with a simple sampling-based method
that conditions on our keypoint representation through its reward function.
In this way, we can take advantage of a constrained prediction space
(of points on the surface of the object)
while still generating motion in the full action space
(so that our model is free to discover new ways of using tools).
Our approach quantitatively matches the performance of a human oracle on two of three tasks
and beats heuristic baselines on all three.
Future work could extend this keypoint representation with additional property encodings
(e.g. for materials)
or integrate object reconstruction
such that plans made with our learned representation in a simulator
can be executed on real robots.

%% file: 7-appendices.tex
\section*{Appendix A: Training Details}

\begin{figure*}[t]
\centering
\includegraphics[width=0.9\linewidth]{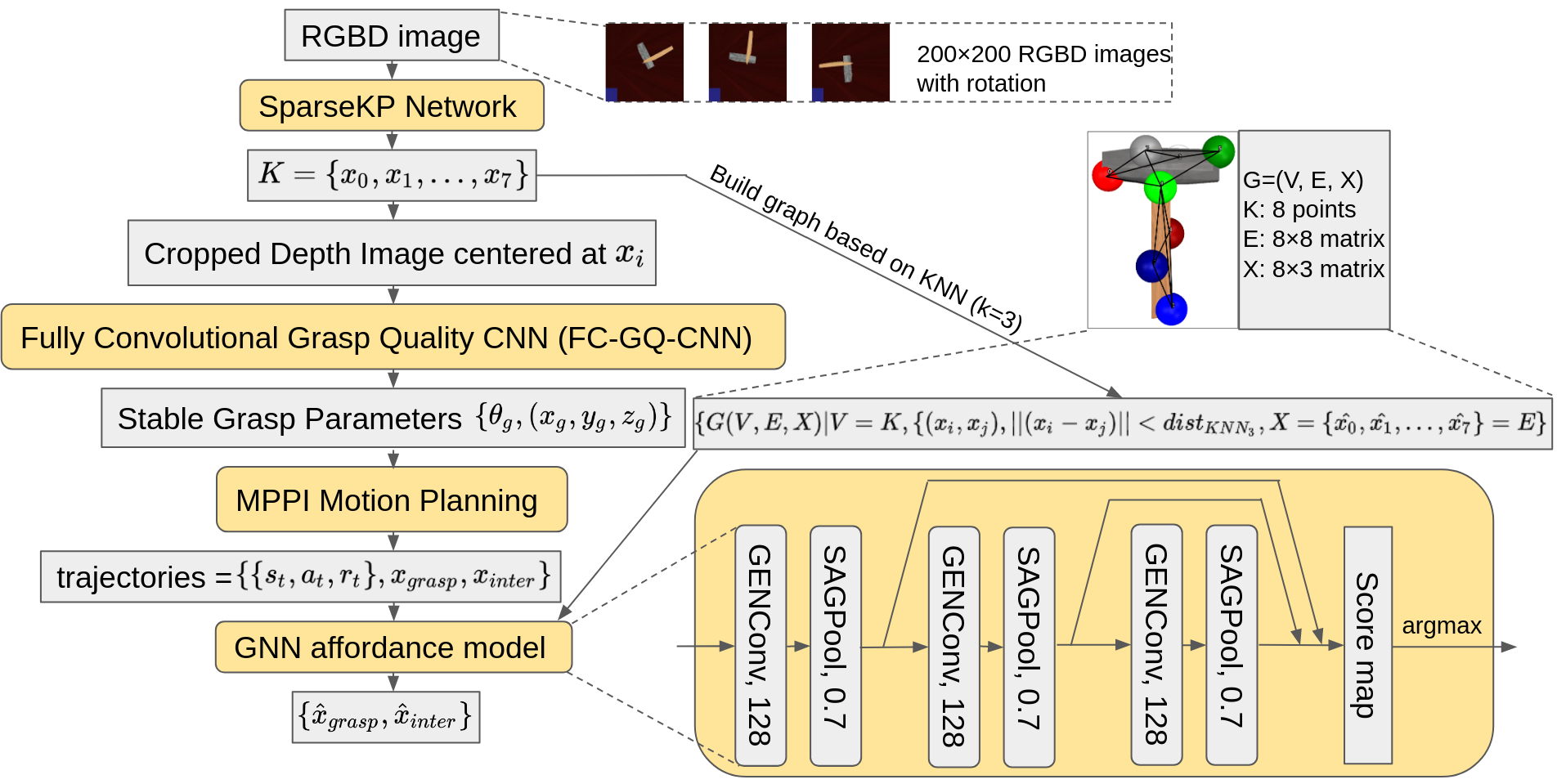}
 \captionof{figure}{\textbf{Network Architecture.}
 A more detailed view of our pipeline.
 The pipeline predicts a grasp keypoint and interaction keypoint from an RGBD image of a tool.
 The GNN affordance model uses residual network layers, graph convolution layers and self-attention graph pooling layers.
}\label{fig:network}
\end{figure*}

\begin{figure}
\centering
\includegraphics[width=\linewidth]{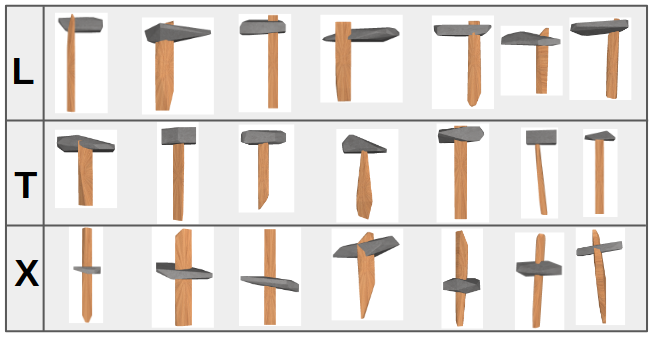}
 \captionof{figure}{\textbf{Procedural Tool Shapes.} We generate tools as a concatenation of two convex shapes in L, T and X configurations.
}\label{fig:tools-ltx}
\end{figure}

\textbf{Tool Datasets.}
We generate a total of 10K tools following the procedure from Section~\ref{sec:experiments:dataset}.
We use all 10K tools when generating training data for the SparseKP Network.
For the affordance models, we select a training set of 300 tools and a non-overlapping evaluation set of 100 tools.
The training and evaluation sets are the same across all three manipulation tasks.
We use tools from the training set when sampling trajectories, and extract affordance instances from said trajectories to train the affordance models.
We use the evaluation set when sampling trajectories to evaluate model performance.
All reward and success numbers in Table~\ref{tab:my-table} are based on trajectories sampled using the evaluation tools.
In Tables~\ref{tab:t-shape}, \ref{tab:l-shape} and \ref{tab:x-shape}, we present results of an evaluation on three additional non-overlapping sets of T-shaped, L-shaped and X-shaped tools.
Each of the shape-specific sets contains 100 tools.

%\paragraph{SparseKP Network.}
\textbf{SparseKP Network.}
We train the SparseKP Network to output keypoints that represent an object's geometry and provide options of regions of potential grasp and interaction during the tool-use.
It is worth noting that this sparse keypoint-based representation is task agnostic. Hence, we train the SparseKP network once and freeze its weights for the rest of the procedure before sampling any trajectories or training the GNN affordance models.
We train SparseKP on a set of 1 million $200\times200$ RGBD tool images.
To generate these images, we use all of the 10K generated tools.
For each tool, we generate 100 poses with uniformly varying $z$-axis angles between $0$ and $2\pi$.
For each pose, we place the tool in the specified pose on a table in a MuJoCo simulator environment, similar to the task environments, and capture RGBD images from a camera positioned above the table with a top-down perspective.
To train the network, we use the ADAM optimizer~\cite{kingma2014adam} with a learning rate of $5\cdot10^{-5}$ and train for $15$ epochs.
Our network architecture is inspired by 6-PACK~\cite{wang20206}, but we make a different choice of loss functions which are given in Section~\ref{sec:method:keypoints}.

%In SparseKP network, we use the same network architecture as 6pack but apply this model to our new reward to detect geometric representable keypoints. Since the SparseKP network is a generic model that not depends on any tasks, we decide to train it separately. At the pretraining stage, the input is a set of RGBD images. For each unseen objects, we manually rotate it at an arbitrary angle at the z-axis and translate a tiny distance. Then we take the RGBD image to such a specific pose. For each unseen object, we have created 100 poses which refers that for a single object, there are 100 RGBD images with different angles and translation parameters. We collected 1k objects RGBD images (which is 1000k RGBD images in total) to train the SparseKP model.

\textbf{Task initialization}
consists of setting starting conditions before our trajectory sampling procedure begins.
First, we uniformly sample a tool -- from the training set during training or the evaluation set during testing.
Next, we uniformly sample a $z$-axis angle from  between $0$ and $2\pi$.
We place the selected tool at the sampled $z$-axis angle positioned $0.1$ units above the table.
We then step the simulator forward, allowing the tool to fall, so that it is resting on the table when the task begins.

\textbf{Trajectory sampling.}
We train a separate affordance model for each task.
To train each model, we sample 10K trajectories.
To sample a trajectory, we start by initializing the environment, as explained above.
Next, we take an observation to infer keypoints from.
We capture a $200\times200$ RGBD image from a camera positioned above the table with a top-down view.
This is passed to our SparseKP network which outputs a set $K$ of 8 task-agnostic keypoints.
To choose a grasping area, we uniformly sample a grasp keypoint $\kp{grasp}$ from $K$.

For grasping, we take a second RGBD observation from the same camera, this time with a resolution of $500\times500$.
Since we only want to consider grasps near $\kp{grasp}$, we take a $200\times200$ crop of the grasping observation centered around $\kp{grasp}$.
We normalize this depth image to match the range of the dataset that our DexNet model was trained on.
We pass this crop to DexNet which produces 50 grasps proposals, each with grasp parameters (position and $z$-axis angle) and grasp quality metric (Q-value).
We select the grasp with position closest to $\kp{grasp}$ with Q-value greater than $0.9$.
If no grasp has a Q-value greater than $0.9$, we take the first grasp.
To execute the grasp, we set the manipulation action based on the position and angle of the grasp.
During training, we only proceed to sample the rest of the trajectory if the grasp is stable.
To check stability, we rotate the gripper 90 degrees in both clockwise and counter-clockwise direction.
We proceed to motion planning only if the hammer is still in grasp after this adversarial rotation test.

\textbf{Affordance Model Training.}
For each of our three tasks, we sample 10K trajectories using uniformly sampled grasping and interaction keypoints.
After extracting the tuples $(K, \kp{grasp}, \kp{inter}, R)$ that constitute the dataset, we train the model using REINFORCE as described in Section~\ref{sec:method:affmodel}.
We use the ADAM optimizer~\cite{kingma2014adam} with a learning rate of $3\cdot10^{-4}$ and train for $700$ epochs.

\textbf{Evaluation.}
For the evaluation reported in the Table~\ref{tab:my-table}, we use a set of 100 tools that were not used in any training trajectories.
Using the same pipeline, we generate 4K trajectories using these evaluation tools.
During \emph{data generation}, we choose grasp and interaction keypoints by uniform sampling.
During \emph{evaluation}, on the other hand, we select grasp and interaction keypoints greedily based on the output of the affordance model (see Figure~\ref{fig:network}).
We present results from these 4K trajectories in Table~\ref{tab:my-table}.

In Tables~\ref{tab:t-shape}, \ref{tab:l-shape} and \ref{tab:x-shape}, we present additional results from trajectories using T, L, and X-shaped tools.
The results were from a mix of these shapes.

\section*{Appendix B: Task Definitions}
\textbf{Hooking Task.}
The setup of the hooking task is shown in Figure~\ref{fig:hooking}.
The goal is to pull the thermos to the goal position (the green point in the figure) by using a tool to hook the handle.
The thermos can only move in the $xy$ plane.
It cannot be picked up.
Performance on  this task is measured by how close the thermos is to the goal position.
The thermos starts out in front of the goal position, which means that ideally, the thermos will move only along the y-axis.
The complete reward $R_{hook}$ is the sum of two terms.
The first term $C_{hook}$ encourages hooking the thermos.
The second term encourages bringing the provisional interaction keypoint $\hat{\kp{inter}}$ towards the target position $x_{target}$ which is the center of the thermos handle.
The formulation of the reward function is as follows:
\begin{equation}
    R_{hook} = w_{hook} C_{hook} - \tanh{(\norm{\hat{\kp{}}_{inter} - x_{target}}_2)}
\end{equation}

\begin{equation}
    C_{hook} = \begin{cases}
        1 - \tanh(\norm{x_{thermos}-x_{goal}}_2) &\parbox[t]{2.2 cm}{if touching and no penalty,}\\[0.5cm]
        0 & \text{otherwise}\\
        \end{cases}
\end{equation}
where $x_{thermos}$ refers to the position of the thermos on the $xy$ plane,
$x_{goal}$ refers to the goal position on the $xy$ plane
and $x_{target}$ is defined as a point on the center of the handle.
We set $w_{hook} = 10,000$.
The touching condition is true if the tool is touching the handle.
The penalty condition is true in 4 situations: 1) the gripper drops the tool; 2) the gripper touches the thermos; 3) the tool contacts the thermos body instead of the handle; and 4) the thermos moves too far on the x-axis since it is supposed to move along the y-axis only.

\textbf{Reaching Task.}
The setup of the reaching task is shown in Figure~\ref{fig:reaching}.
A wall stands between the robot arm and a cylinder, which can only be reached through a small hole in the wall.
The goal is to use a tool to reach through this hole and push the cylinder.
Task performance is measured by how far the cylinder is pushed.

The complete reward $R_{reach}$ is the sum of two terms.
The first term $C_{reach}$ encourages moving the cylinder.
The second term encourages bringing the provisional interaction keypoint $\hat{\kp{inter}}$ towards the target position $x_{target}$.
The formulation of the reward function is as follows:
\begin{equation}
    R_{reach} = w_{reach} C_{reach} - \tanh{(\norm{\hat{\kp{inter}} - x_{target}}_2)},
\end{equation}

\begin{equation}
    C_{reach} = \begin{cases}
        \tanh{(\norm{x_{cylinder}-x_{init}}_2)} &\parbox[t]{2.5cm}{if reaching and no penalty,}\\[0.5cm]
        0 & \text{otherwise}\\
        \end{cases}
\end{equation}
where $x_{cylinder}$ refers to the current position of the cylinder on the $xy$ plane,
$x_{init}$ refers to the initial position of the cylinder
and $x_{target}$ is defined as a point at the entrance of the hole.
The reaching condition is true if the tool is near $x_{target}$.
The penalty condition is true if the gripper drops the tool or the gripper touches the wall.
We set $w_{reach}=10,000$.

\textbf{Hammering Task.}
The setup of the hammering task is shown in Figure~\ref{fig:hammering}.
The goal is to hit the peg with the tool in order to drive the peg into the box.
To encourage swinging instead of pushing, reward for this task is based on the instantaneous acceleration of the peg in the direction of the box when the peg is first touched by the tool.
To avoid solutions where the tool is not used at all, reward is set to $0$ if the gripper touches the peg.
The full reward is given by
\begin{align}
    R_{hammer} &= w_{hammer} C_{hammer} - \tanh{(\norm{\hat{\kp{}}_{inter} - x_{target}}_2)},\\
    C_{hammer} &= \begin{cases}
        (\overrightarrow{a_{peg}}^{goal})^2 &\parbox[t]{3.5cm}{if first contact and no penalty,}\\[0.5cm]
        0 & \text{otherwise}
        \end{cases}
\end{align}
where $\overrightarrow{a_{peg}}^{goal}$ is the component of the acceleration of the peg that is in the direction of the goal.
The first contact condition is true if the tool is touching the peg for the first time.
The penalty condition is true if the tool is dropped or the gripper is in contact with the peg.
We set $w_{hammer}=1$, since peg acceleration is large.

\input{7-t-shapes-table}
\input{7-l-shapes-table}
\input{7-x-shapes-table}

%% file: 7-t-shapes-table.tex
% \usepackage{booktabs}
% \usepackage{graphicx}
\begin{table*}[t!]
\centering
\resizebox{\textwidth}{!}{%
\begin{tabular}{@{}lllrrrrrrr@{}}
\toprule
\multicolumn{1}{l|}{} &  & \multicolumn{1}{c|}{} & \multicolumn{3}{c|}{\textbf{Overall Performance}} & \multicolumn{4}{c}{\textbf{Performance (conditioned on Grasp Success)}} \\ \cmidrule(l){4-10} 
\multicolumn{1}{l|}{\multirow{-2}{*}{}} & \multirow{-2}{*}{\textbf{Grasping Method}} & \multicolumn{1}{c|}{\multirow{-2}{*}{\textbf{Interaction Method}}} & \multicolumn{1}{c|}{\textbf{\begin{tabular}[c]{@{}c@{}}Task \\ Success (\%)\end{tabular}}} & \multicolumn{1}{c|}{\textbf{\begin{tabular}[c]{@{}c@{}}Absolute \\ Reward\end{tabular}}} & \multicolumn{1}{c|}{\textbf{\begin{tabular}[c]{@{}c@{}}Normalized \\ Reward\end{tabular}}} & \multicolumn{1}{c|}{\textbf{\begin{tabular}[c]{@{}c@{}}Grasp \\ Success (\%)\end{tabular}}} & \multicolumn{1}{c|}{\textbf{\begin{tabular}[c]{@{}c@{}}Task \\ Success (\%)\end{tabular}}} & \multicolumn{1}{c|}{\textbf{\begin{tabular}[c]{@{}c@{}}Absolute \\ Reward\end{tabular}}} & \multicolumn{1}{c}{\textbf{\begin{tabular}[c]{@{}c@{}}Normalized \\ Reward\end{tabular}}} \\ \midrule
\multicolumn{10}{l}{\textbf{Hooking}} \\ \midrule
\multicolumn{1}{l|}{Simple baseline} & Antipodal & \multicolumn{1}{l|}{Random (sparseKP)} &33.1\% & 519.32 & \multicolumn{1}{r|}{0.536} & 51.2\% & 64.7\% & 914.31 & 0.706 \\
\multicolumn{1}{l|}{Grasp Optimized baseline} & Stability (DexNet) & \multicolumn{1}{l|}{Random (sparseKP)} & 41.2\% & 716.19 & \multicolumn{1}{r|}{0.739} & 71.2\% & 57.8\% & 994.10 & 0.767 \\
\multicolumn{1}{l|}{Leverage Heuristic baseline} & Stability (DexNet) & \multicolumn{1}{l|}{Farthest (sparseKP)} & 49.2\% & 846.19 & \multicolumn{1}{r|}{0.873} & 69.1\% & 71.2\% & 1,198.97 & 0.926 \\
\multicolumn{1}{l|}{\algoName (Ours)} & \algoName & \multicolumn{1}{l|}{\algoName} & \textbf{55.2\%} & \textbf{949.91} & \multicolumn{1}{r|}{\textbf{0.980}} & 74.2\% & \textbf{74.4\%} & \textbf{1,201.09} & \textbf{0.927} \\ \midrule
\multicolumn{10}{l}{\textbf{Reaching}} \\ \midrule
\multicolumn{1}{l|}{Simple baseline} & Antipodal & \multicolumn{1}{l|}{Random (sparseKP)} & 41.2\% & 401.91 & \multicolumn{1}{r|}{0.769} & 60.1\% & 68.4\% & 664.16 & 0.881 \\
\multicolumn{1}{l|}{Grasp Optimized baseline} & Stability (DexNet) & \multicolumn{1}{l|}{Random (sparseKP)} & 55.9\% & 456.71 & \multicolumn{1}{r|}{0.874} & 63.9\% & 87.5\% & 716.42 & 0.950 \\
\multicolumn{1}{l|}{Leverage Heuristic baseline} & Stability (DexNet) & \multicolumn{1}{l|}{Farthest (sparseKP)} & 56.2\% & 501.46 & \multicolumn{1}{r|}{0.960} & 68.0\% & 82.6\% & 735.60 & 0.976 \\
\multicolumn{1}{l|}{\algoName (Ours)} & \algoName & \multicolumn{1}{l|}{\algoName} & \textbf{61.9\%} & \textbf{524.94} & \multicolumn{1}{r|}{\textbf{1.005}} & 69.6\% & \textbf{88.9\%} & \textbf{751.97} & \textbf{0.997} \\ \midrule
\multicolumn{10}{l}{\textbf{Hammering}} \\ \midrule
\multicolumn{1}{l|}{Simple baseline} & Antipodal & \multicolumn{1}{l|}{Random (sparseKP)} & 49.4\% & 6.21E+6 & \multicolumn{1}{r|}{0.396} & 58.8\% & 84.0\% & 10.68E+6 & 0.542 \\
\multicolumn{1}{l|}{Grasp Optimized baseline} & Stability (DexNet) & \multicolumn{1}{l|}{Random (sparseKP)} & 65.3\% & 8.12E+6 & \multicolumn{1}{r|}{0.518} & 74.6\% & 87.5\% & 10.96E+6 & 0.556 \\
\multicolumn{1}{l|}{Leverage Heuristic baseline} & Stability (DexNet) & \multicolumn{1}{l|}{Farthest (sparseKP)} & \textbf{71.2\%} & 12.07E+6 & \multicolumn{1}{r|}{0.770} & 74.4\% & 95.7\% & 16.28E+6 & 0.827 \\
\multicolumn{1}{l|}{\algoName (Ours)} & \algoName & \multicolumn{1}{l|}{\algoName} & 68.8\% & \textbf{12.74E+6} & \multicolumn{1}{r|}{\textbf{0.813}} & 74.7\% & \textbf{92.2\%} & \textbf{17.22E+6} & \textbf{0.874} \\ \bottomrule
\end{tabular}%
}
\caption{\textbf{T-shaped tools.} \algoName performance evaluated on a set of $100$ T-shaped tools. Since human oracle choices were not collected for this set, normalized reward is based on human oracle performance on the main evaluation set tools (shown in Table 1 of the main paper).
}
\label{tab:t-shape}
\end{table*}

%% file: 7-l-shapes-table.tex
\begin{table*}[t!]
\centering
\resizebox{\textwidth}{!}{%
\begin{tabular}{@{}lllrrrrrrr@{}}
\toprule
\multicolumn{1}{l|}{} &  & \multicolumn{1}{c|}{} & \multicolumn{3}{c|}{\textbf{Overall Performance}} & \multicolumn{4}{c}{\textbf{Performance (conditioned on Grasp Success)}} \\ \cmidrule(l){4-10} 
\multicolumn{1}{l|}{\multirow{-2}{*}{}} & \multirow{-2}{*}{\textbf{Grasping Method}} & \multicolumn{1}{c|}{\multirow{-2}{*}{\textbf{Interaction Method}}} & \multicolumn{1}{c|}{\textbf{\begin{tabular}[c]{@{}c@{}}Task \\ Success (\%)\end{tabular}}} & \multicolumn{1}{c|}{\textbf{\begin{tabular}[c]{@{}c@{}}Absolute \\ Reward\end{tabular}}} & \multicolumn{1}{c|}{\textbf{\begin{tabular}[c]{@{}c@{}}Normalized \\ Reward\end{tabular}}} & \multicolumn{1}{c|}{\textbf{\begin{tabular}[c]{@{}c@{}}Grasp \\ Success (\%)\end{tabular}}} & \multicolumn{1}{c|}{\textbf{\begin{tabular}[c]{@{}c@{}}Task \\ Success (\%)\end{tabular}}} & \multicolumn{1}{c|}{\textbf{\begin{tabular}[c]{@{}c@{}}Absolute \\ Reward\end{tabular}}} & \multicolumn{1}{c}{\textbf{\begin{tabular}[c]{@{}c@{}}Normalized \\ Reward\end{tabular}}} \\ \midrule
\multicolumn{10}{l}{\textbf{Hooking}} \\ \midrule
\multicolumn{1}{l|}{Simple baseline} & Antipodal & \multicolumn{1}{l|}{Random (sparseKP)} & 33.2\% & 554.12 & \multicolumn{1}{r|}{0.572} & 55.2\% & 60.2\% & 998.24 & 0.771 \\
\multicolumn{1}{l|}{Grasp Optimized baseline} & Stability (DexNet) & \multicolumn{1}{l|}{Random (sparseKP)} & 41.15\% & 796.21 & \multicolumn{1}{r|}{0.821} & 70.5\% & 58.3\% & 1,064.12 & 0.821 \\
\multicolumn{1}{l|}{Leverage Heuristic baseline} & Stability (DexNet) & \multicolumn{1}{l|}{Farthest (sparseKP)} & 55.1\% & 889.16 & \multicolumn{1}{r|}{0.917} & 71.2\% & 77.5\% & \textbf{1,247.16} & \textbf{0.963} \\
\multicolumn{1}{l|}{\algoName (Ours)} & \algoName & \multicolumn{1}{l|}{\algoName} & \textbf{60.1\%} & \textbf{975.16} & \multicolumn{1}{r|}{\textbf{1.006}} & 75.1\% & \textbf{80.0\%} & 1,241.16 & 0.958 \\ \midrule
\multicolumn{10}{l}{\textbf{Reaching}} \\ \midrule
\multicolumn{1}{l|}{Simple baseline} & Antipodal & \multicolumn{1}{l|}{Random (sparseKP)} & 45.6\% & 341.31 & \multicolumn{1}{r|}{0.653} & 55.4\% & 82.3\% & 645.01 & 0.855 \\
\multicolumn{1}{l|}{Grasp Optimized baseline} & Stability (DexNet) & \multicolumn{1}{l|}{Random (sparseKP)} & 55.0\% & 458.61 & \multicolumn{1}{r|}{0.878} & 63.9\% & 86.0\% & 702.10 & 0.931 \\
\multicolumn{1}{l|}{Leverage Heuristic baseline} & Stability (DexNet) & \multicolumn{1}{l|}{Farthest (sparseKP)} & 53.6\% & 503.74 & \multicolumn{1}{r|}{0.964} & 65.7\% & 81.6\% & \textbf{742.13} & \textbf{0.984} \\
\multicolumn{1}{l|}{\algoName (Ours)} & \algoName & \multicolumn{1}{l|}{\algoName} & \textbf{61.3\%} & \textbf{511.82} & \multicolumn{1}{r|}{\textbf{0.980}} & 69.1\% & \textbf{88.7\%} & 741.16 & 0.983 \\ \midrule
\multicolumn{10}{l}{\textbf{Hammering}} \\ \midrule
\multicolumn{1}{l|}{Simple baseline} & Antipodal & \multicolumn{1}{l|}{Random (sparseKP)} & 49.3\% & 6.56E+6 & \multicolumn{1}{r|}{0.418} & 59.1\% & 83.5\% & 11.16E+6 & 0.567 \\
\multicolumn{1}{l|}{Grasp Optimized baseline} & Stability (DexNet) & \multicolumn{1}{l|}{Random (sparseKP)} & 65.4\% & 8.46E+6 & \multicolumn{1}{r|}{0.540} & 74.3\% & 88.0\% & 11.46E+6 & 0.582 \\
\multicolumn{1}{l|}{Leverage Heuristic baseline} & Stability (DexNet) & \multicolumn{1}{l|}{Farthest (sparseKP)} & \textbf{70.3\%} & 12.58E+6 & \multicolumn{1}{r|}{0.803} & 75.1\% & \textbf{93.6}\% & 16.86E+6 & 0.856 \\
\multicolumn{1}{l|}{\algoName (Ours)} & \algoName & \multicolumn{1}{l|}{\algoName} & 67.3\% & \textbf{12.72E+6} & \multicolumn{1}{r|}{\textbf{0.812}} & 73.0\% & 92.1\% & \textbf{17.67E+6} & \textbf{0.897} \\ \bottomrule
\end{tabular}%
}
\caption{\textbf{L-shaped tools.} \algoName performance evaluated on a set of $100$ L-shaped tools. Since human oracle choices were not collected for this set, normalized reward is based on human oracle performance on the main evaluation set tools (shown in Table 1 of the main paper).
}
\label{tab:l-shape}
\end{table*}

%% file: 7-x-shapes-table.tex
\begin{table*}[t!]
\centering
\resizebox{\textwidth}{!}{%
\begin{tabular}{@{}lllrrrrrrr@{}}
\toprule
\multicolumn{1}{l|}{} &  & \multicolumn{1}{c|}{} & \multicolumn{3}{c|}{\textbf{Overall Performance}} & \multicolumn{4}{c}{\textbf{Performance (conditioned on Grasp Success)}} \\ \cmidrule(l){4-10} 
\multicolumn{1}{l|}{\multirow{-2}{*}{}} & \multirow{-2}{*}{\textbf{Grasping Method}} & \multicolumn{1}{c|}{\multirow{-2}{*}{\textbf{Interaction Method}}} & \multicolumn{1}{c|}{\textbf{\begin{tabular}[c]{@{}c@{}}Task \\ Success (\%)\end{tabular}}} & \multicolumn{1}{c|}{\textbf{\begin{tabular}[c]{@{}c@{}}Absolute \\ Reward\end{tabular}}} & \multicolumn{1}{c|}{\textbf{\begin{tabular}[c]{@{}c@{}}Normalized \\ Reward\end{tabular}}} & \multicolumn{1}{c|}{\textbf{\begin{tabular}[c]{@{}c@{}}Grasp \\ Success (\%)\end{tabular}}} & \multicolumn{1}{c|}{\textbf{\begin{tabular}[c]{@{}c@{}}Task \\ Success (\%)\end{tabular}}} & \multicolumn{1}{c|}{\textbf{\begin{tabular}[c]{@{}c@{}}Absolute \\ Reward\end{tabular}}} & \multicolumn{1}{c}{\textbf{\begin{tabular}[c]{@{}c@{}}Normalized \\ Reward\end{tabular}}} \\ \midrule
\multicolumn{10}{l}{\textbf{Hooking}} \\ \midrule
\multicolumn{1}{l|}{Simple baseline} & Antipodal & \multicolumn{1}{l|}{Random (sparseKP)} & 29.3\% & 512.25 & \multicolumn{1}{r|}{0.528} & 50.3\% & 58.3\% & 911.27 & 0.703 \\
\multicolumn{1}{l|}{Grasp Optimized baseline} & Stability (DexNet) & \multicolumn{1}{l|}{Random (sparseKP)} & 39.3\% & 721.35 & \multicolumn{1}{r|}{0.744} & 68.4\% & 57.4\% & 1,002.21 & 0.774 \\
\multicolumn{1}{l|}{Leverage Heuristic baseline} & Stability (DexNet) & \multicolumn{1}{l|}{Farthest (sparseKP)} & 50.2\% & 885.29 & \multicolumn{1}{r|}{0.882} & 70.0\% & 71.7\% & \textbf{1,245.20} & \textbf{0.961} \\
\multicolumn{1}{l|}{\algoName (Ours)} & \algoName & \multicolumn{1}{l|}{\algoName} & \textbf{55.9\%} & \textbf{915.21} & \multicolumn{1}{r|}{\textbf{0.944}} & 74.3\% & \textbf{75.2\%} & 1,215.90 & 0.939 \\ \midrule
\multicolumn{10}{l}{\textbf{Reaching}} \\ \midrule
\multicolumn{1}{l|}{Simple baseline} & Antipodal & \multicolumn{1}{l|}{Random (sparseKP)} & 36.2\% & 291.16 & \multicolumn{1}{r|}{0.557} & 49.2\% & 73.6\% & 610.16 & 0.809 \\
\multicolumn{1}{l|}{Grasp Optimized baseline} & Stability (DexNet) & \multicolumn{1}{l|}{Random (sparseKP)} & 51.2\% & 397.61 & \multicolumn{1}{r|}{0.761} & 63.0\% & 81.2\% & 694.67 & 0.921 \\
\multicolumn{1}{l|}{Leverage Heuristic baseline} & Stability (DexNet) & \multicolumn{1}{l|}{Farthest (sparseKP)} & 55.9\% & 481.27 & \multicolumn{1}{r|}{0.921} & 61.9\% & \textbf{90.3\%} & \textbf{716.97} & \textbf{0.951} \\
\multicolumn{1}{l|}{\algoName (Ours)} & \algoName & \multicolumn{1}{l|}{\algoName} & \textbf{57.8\%} & \textbf{500.19} & \multicolumn{1}{r|}{\textbf{0.958}} & 65.2\% & 88.7\% & 700.98 & 0.930 \\ \midrule
\multicolumn{10}{l}{\textbf{Hammering}} \\ \midrule
\multicolumn{1}{l|}{Simple baseline} & Antipodal & \multicolumn{1}{l|}{Random (sparseKP)} & 28.7\% & 3.76E+6 & \multicolumn{1}{r|}{0.240} & 36.3\% & 79.0\% & 10.53E+6 & 0.535 \\
\multicolumn{1}{l|}{Grasp Optimized baseline} & Stability (DexNet) & \multicolumn{1}{l|}{Random (sparseKP)} & 50.7\% & 6.74E+6 & \multicolumn{1}{r|}{0.430} & 58.9\% & 86.0\% & 11.48E+6 & 0.583 \\
\multicolumn{1}{l|}{Leverage Heuristic baseline} & Stability (DexNet) & \multicolumn{1}{l|}{Farthest (sparseKP)} & \textbf{55.7\%} & 8.96E+6 & \multicolumn{1}{r|}{0.572} & 60.6\% & 91.9\% & 15.03E+6 & 0.763 \\
\multicolumn{1}{l|}{\algoName (Ours)} & \algoName & \multicolumn{1}{l|}{\algoName} & 53.6\% & \textbf{9.30E+6} & \multicolumn{1}{r|}{\textbf{0.594}} & 57.8\% & \textbf{92.8\%} & \textbf{16.20E+6} & \textbf{0.823} \\ \bottomrule
\end{tabular}%
}
\caption{\textbf{X-shaped tools.} \algoName performance evaluated on a set of $100$ X-shaped tools. Since human oracle choices were not collected for this set, normalized reward is based on human oracle performance on the main evaluation set tools (shown in Table 1 of the main paper).
}
\label{tab:x-shape}
\end{table*}